\documentclass[journal]{IEEEtran}
\usepackage{amssymb}
\usepackage{booktabs}
\usepackage{amsmath}
\usepackage{subcaption}
\usepackage{adjustbox}
\usepackage{multirow}
\usepackage{graphicx}

\usepackage{algorithm}    
\usepackage{algorithmic}  

\begin{document}
	
	\title{Scalable Multi-Objective Optimization for Robust Traffic Signal Control in Uncertain Environments}

\author{Weian Guo, Wuzhao Li, Zhiou Zhang, Lun Zhang, Li Li and Dongyang Li
	\thanks{This work is supported by the National Natural Science Foundation of China under Grant Number 62273263, 72171172 and 71771176; Shanghai Municipal Science and Technology Major Project (2022-5-YB-09); Natural Science Foundation of Shanghai under Grant Number 23ZR1465400. (Corresponding author: Li Li). Thanks for Mr. Wanli CAI for his helps in providing experiment environments to run and debug the algorithms. }
	\thanks{Weian GUO and Dongyang LI are with the Sino-German College of Applied Sciences, Tongji University, Shanghai, China. Email:\{guoweian@tongji.edu.cn; lidongyang0412@163.com\}. Wuzhao LI is with the Yongjia College, Wenzhou Polytechnic, Wenzhou, Zhejiang, China. Email:\{lwzhao055@wzpt.edu.cn\}. Zhiou ZHANG is with the School of Computer Science and Engineering, University of New South Wales, Sydney, Australia. Email:\{zhiou.zhang98@gmail.com\}. Lun ZHANG is with the Department of Transportation, Tongji University, Shanghai, China. Email:\{lun\_zhang@tongji.edu.cn\}. Li LI is with the Department of Electrics and Information Engineering, Tongji University, Shanghai, China. Email:\{lili@tongji.edu.cn\}.}
}

\maketitle

\begin{abstract}
	Intelligent traffic signal control is essential to modern urban management, with important impacts on economic efficiency, environmental sustainability, and quality of daily life. However, in current decades, it continues to pose significant challenges in managing large-scale traffic networks, coordinating intersections, and ensuring robustness under uncertain traffic conditions. This paper presents a scalable multi-objective optimization approach for robust traffic signal control in dynamic and uncertain urban environments. A multi-objective optimization model is proposed in this paper, which incorporates stochastic variables and probabilistic traffic patterns to capture traffic flow dynamics and uncertainty. We propose an algorithm named Adaptive Hybrid Multi-Objective Optimization Algorithm (AHMOA), which addresses the uncertainties of city traffic, including network-wide signal coordination, fluctuating patterns, and environmental impacts. AHMOA simultaneously optimizes multiple objectives, such as average delay, network stability, and system robustness, while adapting to unpredictable changes in traffic. The algorithm combines evolutionary strategies with an adaptive mechanism to balance exploration and exploitation, and incorporates a memory-based evaluation mechanism to leverage historical traffic data. Simulations are conducted in different cities including Manhattan, Paris, São Paulo, and Istanbul. The experimental results demonstrate that AHMOA consistently outperforms several state-of-the-art algorithms and the algorithm is competent to provide scalable, robust Pareto optimal solutions for managing complex traffic systems under uncertain environments.
\end{abstract}


\section{Introduction}
Intelligent traffic systems are one of the most critical challenges faced by modern cities, impacting economic efficiency, environmental sustainability, and overall quality of life. The inherent complexities of urban traffic systems—characterized by dynamic and unpredictable traffic patterns, multiple conflicting optimization objectives, and stochastic elements—present significant challenges. Effective traffic signal optimization is essential to improving traffic flow, reducing congestion, and lowering greenhouse gas emissions \cite{papageorgiou2003review,li2014efficient}. Traffic signal control is inherently a multi-objective optimization problem involving objectives that often conflict with each other. The challenges caused by expanding scale of urban traffic networks necessitates optimization techniques capable of efficiently and effectively managing large-scale problems. The interdependency among different intersections, combined with varying travel demands, increases the fluctuations of traffic flow and affects overall traffic dynamics. Furthermore, external variables such as meteorological conditions and unforeseen circumstances provide additional levels of unpredictability and intricacy to traffic flow \cite{Ma2020multi, qadri2020state}.

Current research focuses on real-time traffic signal adjustments for dynamic scenarios, considering the interconnections among large-scale urban intersections and the stability of the traffic system. However, frequent strategy switching can impact the overall stability of urban traffic. Therefore, it is crucial to develop robust multi-objective optimization strategies for large-scale urban traffic networks to handle the dynamic and stochastic nature of traffic flow and maintain stable traffic conditions. To address these needs, we conduct comprehensive work in both scenario modeling and algorithm design.

Firstly, in modeling traffic signal scenarios, we incorporate random factors such as weather, holidays, and emergency events that influence traffic flow to varying degrees. To provide a comprehensive perspective for a city, we consider thousands of intersections and the effects of vehicle flow among adjacent intersections. Therefore, we frame the modeling as a large-scale multi-objective optimization problem with inherent uncertainties. Our research addresses a range of complex urban scenarios, including grid-based structures (e.g., Manhattan), radial-concentric layouts (e.g., Paris), rapidly expanding metropolises (e.g., São Paulo), and complex multi-nodal environments (e.g., Istanbul). By considering these diverse urban topologies, we aim to create a robust and adaptable solution capable of handling the multifaceted nature of urban traffic management.

Secondly, for the algorithm design, we propose the Adaptive Hybrid Multi-Objective Optimization Algorithm (AHMOA), which integrates multiple evolutionary strategies and adaptively adjusts their usage probabilities based on their success rates during the optimization process. AHMOA is a comprehensive multi-objective optimization algorithm designed for urban traffic signal control. It employs an adaptive multi-strategy mechanism to dynamically balance exploration and exploitation, thereby efficiently navigating the large-scale search space of urban traffic signal control problems. To enhance robustness, AHMOA incorporates a memory-based evaluation mechanism that accounts for historical traffic patterns and variations. This mechanism improves the stability and reliability of solutions by considering past traffic data, making the optimization process more robust to stochastic fluctuations and external disruptions.

This approach aims to provide a more stable and reliable solution for urban traffic management by effectively addressing the inherent complexities and dynamic nature of urban traffic systems. The main contributions of this work are as follows:

\begin{itemize} 
	\item Proposal of a robust multi-objective optimization model that simultaneously optimizes average delay, network stability, and system robustness, addressing uncertainties in traffic environments and ensuring stable performance under varying conditions.
	\item Development of a large-scale, adaptive multi-objective optimization algorithm (AHMOA) for traffic signal control, capable of managing extensive urban networks and effectively responding to dynamic traffic patterns and unexpected changes in traffic conditions.
	\item Validation and demonstration of the algorithm's versatility through comprehensive simulations in four distinctly different urban settings (Manhattan, Paris, São Paulo, and Istanbul), showcasing its effectiveness across diverse city layouts and its ability to coordinate traffic signals in large urban areas while accounting for intersection dependencies.
\end{itemize}

The remainder of this paper is organized as follows: Section \ref{sec:related} provides a comprehensive review of related work, highlighting recent advancements in traffic signal control and identifying current research gaps. In Section \ref{sec:pro}, we present a detailed formulation of the problem and introduce our proposed optimization framework. Section \ref{sec:method} elucidates the methodology of our novel Adaptive Hybrid Multi-Objective Optimization Algorithm (AHMOA), detailing its key components and operational mechanisms. Extensive experimental results and comparative analyses are presented in Section \ref{sec:exp}, demonstrating the efficacy of our approach across diverse urban environments. Finally, Section \ref{sec:con} concludes the paper with a summary of our key findings, discusses the broader implications of our work, and outlines promising directions for future research in this domain.

\section{Related Work}
\label{sec:related}
Traffic signal control is a critical aspect of urban traffic management aimed at optimizing traffic flow, reducing congestion, and enhancing overall efficiency \cite{shaikh2022review}. This section reviews the evolution of traffic signal control methods, focusing on the developments in optimization algorithms, machine learning and emerging technologies.

\subsection{Traditional and Adaptive Traffic Signal Control}
The evolution of traffic signal control has seen a transition from traditional fixed-time controllers to more adaptive systems. Early systems, such as those described by Webster \cite{webster1958traffic} and Robertson \cite{robertson1969transyt}, relied on historical traffic data to set signal timings. These fixed-time controllers, while simple and reliable, lacked the robustness and flexibility to respond to traffic fluctuations, leading to inefficiencies in dynamic traffic conditions. In response, actuated controllers are developed, adjusting signal timings based on real-time sensor feedback \cite{roess2011traffic}. However, these systems still struggled with optimality and adaptability, particularly in complex urban environments \cite{wang2024traffic}. To address these limitations, adaptive traffic signal control (ATSC) systems have been developed, leveraging real-time traffic data to dynamically adjust signal timings, thereby significantly enhancing traffic flow and reducing delays \cite{shabestary2022adaptive}. Despite their advantages, early ATSC systems faced challenges in handling the complexity and variability of modern urban traffic, especially in large-scale networks.

\subsection{Optimization Techniques in Traffic Signal Control}
Recent advancements have integrated sophisticated optimization algorithms to enhance the adaptability and efficiency of traffic control systems.

\subsubsection{Multi-Objective Optimization}
Multi-objective optimization has been widely adopted to balance conflicting objectives such as minimizing delays, reducing fuel consumption, maximizing throughput and many others. The corresponding algorithms are competent to provide diverse Pareto-optimal solutions to address the traffic signal problems \cite{jin2019multi}. Tan et al. \cite{Tan2024Connected} and Lin et al. \cite{lin2023traffic} demonstrated the integration of real-time data into these models for dynamic optimization in urban environments, showcasing significant potential for real-time traffic management. However, these methods often require extensive computational resources and may not scale well in large scale applications and dynamic traffic environments. To address this, Gao et al. \cite{gao2017jaya} explored metaheuristic approaches, including the Jaya algorithm, harmony search (HS), and water cycle algorithm (WCA), demonstrating their effectiveness in handling complex, large-scale traffic optimization problems. Despite their advantages, the scalability and real-time application of these methods remain areas for further research.

\subsubsection{Machine Learning and Reinforcement Learning}
Machine learning techniques, particularly reinforcement learning (RL), have shown significant promise in developing more adaptive and intelligent traffic signal control systems. Deep reinforcement learning (DRL) approaches, such as those proposed by Gao et al. \cite{gao2021information} and Wei et al. \cite{wei2019intellilight}, have demonstrated substantial improvements by learning optimal policies from traffic data. These methods, however, often face challenges related to the high-dimensional state and action spaces in traffic control scenarios. To mitigate these issues, researchers have integrated advanced sensing technologies and deep learning models. For instance, Shabestary and Abdulhai \cite{shabestary2022adaptive} used GPS traces from connected vehicles, employing Deep Q-Networks (DQN) for processing and optimizing signal timings. Ma et al. \cite{Ma2022deep} and Yang et al. \cite{Yang2024enhancing} further advanced this field by combining computer vision techniques with reinforcement learning, highlighting the potential for improved adaptability and efficiency. Research by Gu et al. \cite{gu2023large} and Chiou \cite{chiou2023cooperative} focused on developing traffic control systems that can adapt to changing conditions and recover from disruptions. 

\subsection{Network-Wide and Multi-Agent Approaches}
The complexity of urban traffic networks has necessitated the development of network-wide and multi-agent approaches to traffic signal control.

\subsubsection{Network-Wide Optimization}
Network-wide optimization methods consider interactions between intersections, using predictive models to adapt signal timings dynamically. Wang et al. \cite{wang2023network} and Yao et al. \cite{yao2020dynamic} leveraged real-time data to optimize traffic flow across entire networks, addressing issues related to large-scale coordination and dynamic traffic conditions. Zhu et al. \cite{zhu2022bi} employed game-theoretic approaches to manage network-wide traffic, demonstrating the potential of such methods in coordinating multiple bottlenecks. These approaches, while effective, often require robust data communication infrastructures and face challenges in scalability and real-time implementation. Future research should focus on developing more efficient algorithms and improving the scalability of these methods.

\subsubsection{Multi-Agent Systems}
Multi-agent systems and graph-based methods have been explored to capture the spatial relationships within traffic networks and enable decentralized control. Wang et al. \cite{wang2022gan} and Devailly et al. \cite{devailly2022ig} demonstrated the effectiveness of multi-agent deep reinforcement learning and graph-convolutional networks in achieving coordinated traffic management. Jiang et al. \cite{Jiang2022distributed} applied graph theory to decompose complex traffic networks for more efficient optimization, highlighting the potential of these approaches in managing large-scale urban traffic. However, the decentralized nature of multi-agent systems can lead to challenges in coordination and consistency across agents. Addressing these challenges through improved communication protocols and coordination algorithms is an area for future exploration.

\subsection{Emerging Technologies and Future Directions}
The integration of emerging technologies such as the Internet of Things, connected vehicles, and smart cities infrastructure offers new opportunities for traffic signal control.

\subsubsection{Connected Vehicle Technologies}
Connected vehicle technologies facilitate precise and coordinated traffic management strategies by enabling communication between vehicles and infrastructure. Research by Emami et al. \cite{emami2022network} and Feng et al. \cite{feng2022cybersecurity} highlighted the potential of integrating connected vehicle data into traffic signal optimization, improving real-time responsiveness and overall system efficiency. These technologies, however, raise concerns regarding data privacy and the need for robust cybersecurity measures.

\subsubsection{IoT and Computer Vision}
IoT devices provide real-time traffic data, enhancing adaptive and predictive capabilities of traffic control systems. Studies by Ferguson et al. \cite{ferguson2020iot} and Hasan et al. \cite{hasan2021iot} demonstrated the effectiveness of IoT in traffic management. Additionally, Chu et al. \cite{chu2022traffic} explored the use of computer vision techniques for real-time image processing at intersections, integrating these methods with reinforcement learning for enhanced traffic control. While promising, the integration of IoT and computer vision technologies requires substantial investment in infrastructure and faces challenges related to data integration and processing capabilities.

\subsection{Summary of Related Work}
While significant advancements have been made in traffic signal control through the integration of optimization algorithms, machine learning techniques, and emerging technologies, several limitations remain. Traditional and adaptive traffic control systems often struggle to handle the complexity and variability of large-scale urban traffic, particularly in dynamic and uncertain environments. Optimization-based methods, though effective in achieving diverse Pareto solutions, frequently face scalability challenges and require extensive computational resources, limiting their real-time application in large-scale scenarios. Furthermore, machine learning approaches, such as reinforcement learning, encounter difficulties in managing high-dimensional state and action spaces, leading to suboptimal results in large traffic networks. The integration of IoT and connected vehicle data shows promise but introduces issues related to data privacy, cybersecurity, and infrastructure costs.

\begin{figure}[!htbp]
	\centering
	\includegraphics[width=0.5\textwidth]{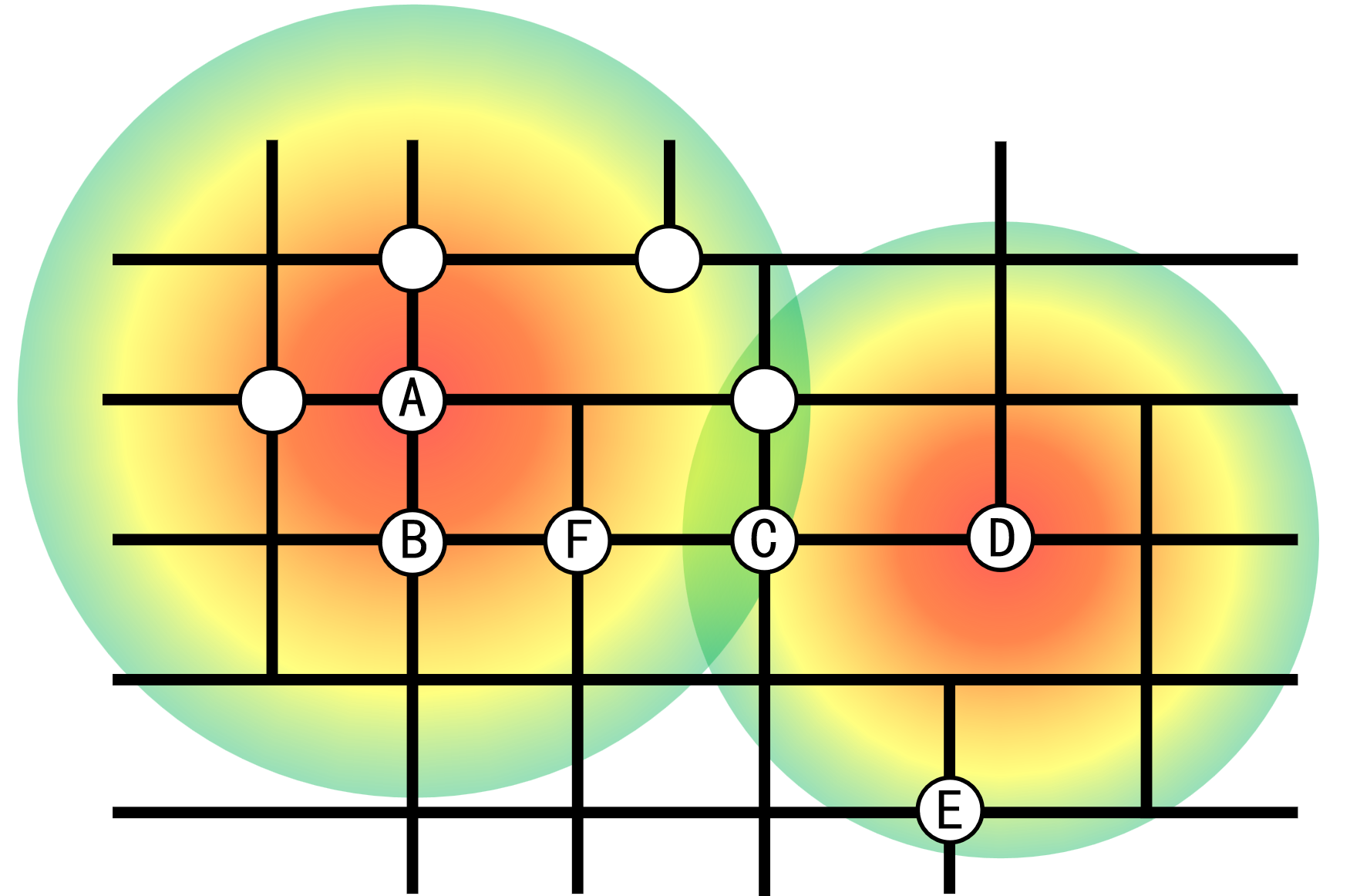}
	\caption{An example of a local area in a large-scale traffic network}
	\label{fig:intersections}
\end{figure}
In large-scale traffic networks, the flow of vehicles between intersections exhibits both direct and indirect propagation. As shown in Fig.~\ref{fig:intersections}, the impact of traffic flow at an intersection radiates outward, with its intensity decreasing from high to low (as indicated by the gradient from red to green). Congestion or relief at point A will directly impact the traffic flow at the next intersection, point B, and indirectly influence the traffic at point F. This effect cascades through the network, potentially affecting more distant intersections such as point E. Moreover, traffic flow propagation is influenced by multiple factors. For instance, the flow at point C is impacted by a combination of flows from both points A and D. As a result, designing an algorithm for controlling traffic signals at an isolated intersection without considering the entire network may lead to sub-optimal outcomes, where improvements at one point come at the expense of others. In addition, external factors such as time of day and weather can further complicate signal control in different regions of the network. Traditional methods, which rely solely on local or even single-intersection traffic data to design traffic signal control strategies, fail to holistically account for the overall traffic conditions across the city. Frequent switching of such strategies can lead to instability in the traffic signal network. Therefore, in the face of uncertain environments, it is of great significance to develop robust large-scale traffic signal control strategies that can effectively manage the entire urban road network.

This study addresses these limitations by proposing the Adaptive Hybrid Multi-Objective Optimization Algorithm (AHMOA), which combines multiple optimization strategies to balance exploration and exploitation while adapting dynamically to the problem’s landscape. AHMOA is designed to scale effectively in large-scale urban traffic networks by incorporating memory-based evaluation, which reduces computational overhead, and by leveraging adaptive strategy selection to optimize traffic flow in real-time. In addition, by considering robustness as a key objective, AHMOA provides solutions that maintain performance across varying traffic conditions, enhancing both scalability and stability in dynamic environments.

\section{Problem Formulation and System Modeling}
\label{sec:pro}
In this section, we establish a robust large-scale multi-objective model for urban traffic signal optimization. 
\subsection{Traffic Environment}
In this model, cycle lengths ($C$) represent the total duration of a traffic signal cycle, including green, yellow, and red phases, directly affecting the delay time at each intersection and the overall traffic flow efficiency. Saturation flow rates ($SR$) denote the maximum rate at which vehicles can pass through an intersection under ideal conditions, used to calculate the intersection's capacity and influence delay and queue length computations. Traffic volumes ($V$) indicate the number of vehicles passing through an intersection over a specified period, essential for evaluating the load on each intersection and affecting both delay and network stability calculations. Adjacency matrices ($AM$) represent the connectivity between intersections, modeling the relationships and dependencies between different intersections, thereby impacting the overall network stability assessment. Intersection types ($IT$) refer to specific configurations of intersections, such as crossroads or T-junctions, influencing traffic flow patterns and considered when evaluating network stability and optimizing signal timings. 

This model incorporates average delay, network stability, and robustness into the optimization process. Additionally, it considers various city-specific traffic patterns and accounts for uncertainties, such as weather changes and emergent events, in traffic conditions. In this paper, we consider a whole city or a large district, resulting in the number of intersections exceeding thousands. Therefore, we do not focus on the control of lanes and traffic directions at a microscopic level but rather study the traffic flow rates at intersections. The types of intersections considered include crossroads, T-junctions, and others, which will be detailed in the subsequent simulation experiments. Let $\lambda_i \in [0,1]$ represent the red light ratio for intersection $i$, where $i = 1, 2, \ldots, N$, and $N$ is the total number of intersections in the network. In our research, the value of $N$ reaches the scale of several thousand. The decision vector is defined as shown in \eqref{eqn:lambda_define}.
\begin{equation}
	\label{eqn:lambda_define}
	\lambda = [\lambda_1, \lambda_2, \ldots, \lambda_N]
\end{equation}
Therefore, the decision variables in the proposed optimization model are the red light ratios ($\lambda$) for each intersection in the network, represented as a vector of length equal to the number of intersections. The traffic control problem is formulated as a multi-objective optimization task. The detailed objectives are provided as follows.

\subsection{Mathematical Formulation for Objectives}
In this paper, we consider average delay and traffic network flow as the primary objectives for the multi-objective optimization problem, which can be mathematically expressed as follows.

\subsubsection{Average Delay}
The first objective function $f_1(\lambda)$ represents the expected average vehicle delay in the network:
\begin{equation}
	\label{eqn:delay}
	f_1(\lambda) = \mathbb{E}\left[\frac{1}{T} \sum_{t=1}^{T} \sum_{i=1}^{N} D_i(t, \lambda_i)\right]
\end{equation}
\noindent where $T$ is the total number of simulation time segments, and $D_i(t, \lambda_i)$ is the delay at intersection $i$ at time $t$, calculated using a modified Webster's formula:
\begin{equation}
	D_i(t, \lambda_i) = \frac{C_i(1-g_i/C_i)^2}{2(1-\lambda_i x_i(t))} + \frac{x_i(t)^2}{2s_i(t)(1-x_i(t))}
\end{equation}
where:
\begin{itemize}
	\item $C_i$ is the cycle length for intersection $i$
	\item $g_i$ is the effective green time, where $g_i = (1-\lambda_i)C_i$
	\item $x_i(t) = \frac{v_{i,t}}{s_i(t)}$ is the degree of saturation (traffic intensity)
	\item $s_i(t)$ is the saturation flow rate, adjusted for dynamic conditions such as weather
	\item $v_{i,t}$ is the volume of vehicles at intersection $i$ at time $t$
\end{itemize}

This objective aims to reduce the overall waiting time experienced by vehicles in the network. The modified Webster's formula accounts for both uniform and random delays, capturing the impact of red light ratios on waiting times and queue lengths.

\subsubsection{Network Stability}
Network stability is crucial for maintaining smooth and predictable traffic flow, which helps in reducing congestion and improving overall traffic efficiency. Therefore, in this paper, the second objective function $f_2(\lambda)$ aims to maximize network stability, which reflects the consistency and predictability of traffic flow across the network:
\begin{align}
	\label{eqn:stability}
	f_2(\lambda) &= \mathbb{E}\Bigg[\sum_{t=1}^T \sum_{i=1}^N \big(|v_{i,t} - v_{i+1,t}| + |v_{i,t} - v_{i,t+1}|\big) \notag \\
	&\qquad \cdot \big(1 + |\lambda_i - \lambda_{i+1}|\big) \cdot AM_{i,i+1} \cdot IT_i \Bigg]
\end{align}
\noindent where $v_{i,t}$ is the vehicle count at intersection $i$ at time $t$, $AM_{i,i+1}$ is the adjacency matrix element indicating if intersections $i$ and $i+1$ are connected, and $IT_i$ represents the type of intersection $i$.

This objective captures the interdependencies between different intersections and the effect of traffic flow variations on network stability. By minimizing the differences in vehicle counts between adjacent intersections ($|v_{i,t} - v_{i+1,t}|$) and between consecutive time intervals ($|v_{i,t} - v_{i,t+1}|$), and considering the variations in red light ratios ($|\lambda_i - \lambda_{i+1}|$), we ensure that traffic signals are coordinated to maintain a balanced and stable traffic network. The inclusion of $AM_{i,i+1}$ allows the model to account for the connectivity between intersections, while $IT_i$ helps to weigh the impact of different intersection types on traffic flow and stability. This coordination is essential for mitigating the impact of sudden changes in traffic flow and ensuring that traffic signal strategies are robust and adaptable to dynamic traffic conditions.

\subsubsection{Robust Objective}
To account for uncertainties in traffic patterns and environmental conditions, we employ a comprehensive robust evaluation approach. To enhance the robustness of our solutions, we incorporate a stability score as an additional measure. This score is calculated for each solution and is used in the Pareto front selection process. The stability score is defined in \eqref{eqn:robust}, which can be considered as the third objective.
\begin{equation}
	\label{eqn:robust}
	R(\lambda) = \frac{1}{M} \sum_{j=1}^{M} \sqrt{\frac{1}{N-1} \sum_{h=1}^{N} (f_j^h(\lambda) - \overline{f_j(\lambda)})^2}
\end{equation}
\noindent where $M$ is the number of objective functions, $N$ is the number of hours considered, $f_j^h(\lambda)$ is the value of the $j$-th objective function for hour $h$, and $\overline{f_j(\lambda)}$ is the mean value of the $j$-th objective function across all hours, calculated by \eqref{eqn:r_aver}.
\begin{equation}
	\label{eqn:r_aver}
	\overline{f_j(\lambda)} = \frac{1}{N} \sum_{h=1}^{N} f_j^h(\lambda)
\end{equation}
$R$ measures the average standard deviation of each objective function's performance across different hours of the day. A lower value of $R$ indicates a more stable and robust solution, as it represents traffic control strategies that are more robust to varying traffic conditions throughout the day.

In summary, by integrating three objectives, namely $f_1$, $f_2$, and $R$, into a unified optimization model, we develop traffic signal strategies that not only minimize delays and enhance stability but also demonstrate robustness against unpredictable traffic and environmental factors. This approach leads to a reliable and efficient urban traffic management system, capable of adapting to the dynamic nature of modern city traffic.

\subsection{Evaluating Uncertainty Considerations}
To incorporate multiple uncertainties to ensure the solutions are robust to varying traffic conditions, we consider two aspects in the modeling.
\subsubsection{Multiple Evaluations}
In practice of this paper, each solution is evaluated $n_e$ times under different conditions. The final objective values are calculated as the expected values over these evaluations:
\begin{equation}
	\bar{f}_k(\lambda) = \mathbb{E}[f_k(\lambda)] = \frac{1}{n_e} \sum_{j=1}^{n_e} f_k^j(\lambda), \quad k \in \{1, 2\}
\end{equation}
This approach allows us to find solutions that perform well across various time intervals in a whole day.

\subsubsection{Dynamic Weather Simulation}
We introduce a time-dependent weather factor $\omega(t)$ to simulate changing environmental conditions throughout the day:
\begin{equation}
	s_i(t) = s_{i,\text{base}} \cdot \omega(t)
\end{equation}
\noindent where $\omega(t) = 0.8 + 0.4 \sin(2\pi t/T)$. This factor affects the saturation flow rate, reflecting the impact of weather on traffic conditions. This equation simulates the periodic changes in weather conditions throughout the day. Specifically, $\omega(t)$ is a time-varying factor that adjusts the saturation flow rate $s_i(t)$. The sine function $\sin(2\pi t/T)$ models the cyclical nature of daily weather patterns. The baseline value is set at 0.8 with a fluctuation range of 0.4, meaning the weather factor ranges between 0.4 and 1.2, reflecting the extent of weather impact on the saturation flow rate.

\section{Methodology}
\label{sec:method}
The main challenges of the model come from addressing the multi-objective optimization while considering the large scale of the problem and its inherent uncertainties. To address these issues, it is essential to design an algorithm that is adequately suitable for exploring the search space. Therefore, we propose a novel approach named the Adaptive Hybrid Multi-Objective Optimization Algorithm (AHMOA). This algorithm employs an adaptive hybrid multi-strategy scheme, which leverages the strengths of different algorithms to perform efficient exploration and exploitation in the high-dimensional decision space. To address the robustness issue, we incorporate robustness indicators as one of the optimization objectives.  The details are provided as follows. 

\subsection{Memory-Based Evaluation Calculation}
We introduce a memory mechanism in the algorithm design to store and utilize historical traffic data, thereby enhancing the robustness and stability of the algorithm. The memory matrix $M$ is computed from historical vehicle counts $V$. Let $V_{i,t}$ be the vehicle count at intersection $i$ at time $t$ for the current evaluation. The memory matrix $M$ is then defined as the average of the past $H$ historical vehicle counts. The elements of $M$ are given by \eqref{eqn:M}.
\begin{equation}
	\label{eqn:M}
	M_{i,t} = \frac{1}{H} \sum_{h=1}^{H} V_{i,t,h}
\end{equation}
\noindent where  $M_{i,t}$ is the averaged vehicle count at intersection $i$ at time $t$, $V_{i,t,h}$ is the vehicle count at intersection $i$ at time $t$ during the $h$-th historical evaluation, $H$ is the number of historical evaluations considered. In the objectives \eqref{eqn:delay} and \eqref{eqn:stability}, the vehicle count $V_{i,t}$ is replaced with $M_{i,t}$ to incorporate the memory-based evaluation mechanism. The memory matrix $M$ plays a critical role by incorporating past vehicle counts into the current evaluation, helping to mitigate the impact of dynamic and uncertain traffic conditions.

\subsection{Adaptive Offspring Generation}
Offspring are generated using an adaptive hybrid strategy that combines Genetic Algorithm (GA), Differential Evolution (DE), Particle Swarm Optimization (PSO), and Local Search (LS). Strategy probabilities are updated based on their success rates in generating better solutions. This adaptive mechanism allows the algorithm to dynamically balance exploration and exploitation, enhancing its ability to navigate the complex, multi-objective landscape of traffic signal optimization.

Let $\mathbf{p}^{\text{old}}$ be the vector of probabilities for each strategy before updating, and $\mathbf{R}_{\text{succ}}$ be the vector of success rates for each strategy. The updated strategy probabilities $\mathbf{p}^{\text{new}}$ are calculated by \eqref{eqn:p_new}.
\begin{equation}
	\label{eqn:p_new}
	\mathbf{p}^{\text{new}} = (1 - \alpha) \mathbf{p}^{\text{old}} + \alpha \mathbf{R}_{\text{succ}}
\end{equation}
where $\alpha$ is the learning rate. The old probabilities and success rates are defined by \eqref{eqn:p_old} and \eqref{eqn:r_succ}.
\begin{equation}
	\label{eqn:p_old}
	\mathbf{p}^{\text{old}} = [p_{\text{GA}}^{\text{old}}, p_{\text{DE}}^{\text{old}}, p_{\text{PSO}}^{\text{old}}, p_{\text{LS}}^{\text{old}}]
\end{equation}
\begin{equation}
	\label{eqn:r_succ}
	\mathbf{R}_{\text{succ}} = [R_{\text{GA}}, R_{\text{DE}}, R_{\text{PSO}}, R_{\text{LS}}]
\end{equation}
where $p_{i}^{\text{old}}$ represents the old probability for strategy $i$ and $R_{i}$ represents the success rate for strategy $i$. Initially,     $\mathbf{p}^{\text{old}} = [0.25, 0.25, 0.25, 0.25]$ and $    \mathbf{R}_{\text{succ}} =[0,0,0,0]$. The success rate for each strategy $R_{i}$ is calculated by \eqref{eqn:r_cal}.
\begin{equation}
	\label{eqn:r_cal}
	R_{i} = \frac{n_{i}^{\text{succ}}}{n_{\text{total}}}
\end{equation}
where $n_{i}^{\text{succ}}$ is the number of offspring generated by strategy $i$ that improve upon their parent solutions, and $n_{\text{total}}$ is the total number of offspring generated.

This approach ensures that strategies that generate more successful offspring have higher probabilities in future generations, allowing the algorithm to dynamically adapt to the problem's landscape. The Genetic Algorithm (GA) utilizes Simulated Binary Crossover (SBX) and polynomial mutation to generate new solutions, where the crossover operation combines two parent solutions based on a distribution index and a random value, and the mutation introduces small adjustments to enhance diversity. Differential Evolution (DE) creates offspring through differential mutation and binomial crossover, using randomly selected individuals and a differential weight to generate variability. Particle Swarm Optimization (PSO) updates particle velocities based on personal and global best positions, employing inertia weight, cognitive and social coefficients, and random values to guide particles towards optimal solutions. Local Search fine-tunes solutions by applying small perturbations, exploring the neighborhood of each solution to enhance local exploitation. The pseudo-codes are given in Algorithm \ref{alg:offspring_generation}.

\begin{algorithm}
	\caption{Adaptive Hybrid Strategy to Generate Offspring}
	\label{alg:offspring_generation}
	\begin{algorithmic}[1]
		\STATE \textbf{Input:} Strategy probabilities $\mathbf{p}$, population $P$, learning rate $\alpha$
		\STATE Initialize $\mathbf{S} = [0, 0, 0, 0]$ and $\mathbf{T} = [0, 0, 0, 0]$
		\FOR{each individual $i$ in population $P$}
		\STATE Select a strategy $s$ using roulette wheel selection based on probabilities $\mathbf{p}$
		\STATE According to $s$, use GA or DE or PSO, or LS to generate offspring $o_i$
		\STATE Evaluate if the offspring $o_i$ generated by strategy $s$ improves over its parent $p_i$
		\IF{$o_i$ improves over $p_i$}
		\STATE $\mathbf{S}[s] = \mathbf{S}[s] + 1$
		\ENDIF
		\STATE $\mathbf{T}[s] = \mathbf{T}[s] + 1$
		\STATE Add offspring $o_i$ to the offspring population $O$
		\ENDFOR
		\STATE Calculate $\mathbf{R}_{\text{succ}}[s] = \frac{\mathbf{S}[s]}{\mathbf{T}[s]}$ for each strategy $s$
		\STATE Update strategy probabilities $\mathbf{p}$ by \eqref{eqn:p_new}
		\STATE Normalize $\mathbf{p}$ so that $\sum \mathbf{p} = 1$
		\STATE \textbf{Return:} Offspring population $O$, updated strategy probabilities $\mathbf{p}$
	\end{algorithmic}
\end{algorithm}

\subsection{Offspring Selection}
The combined parent and offspring populations are evaluated using a non-dominated sorting approach to form Pareto fronts. In this method, individuals are sorted into different fronts based on dominance. One individual dominates another if it is no worse in all objectives and strictly better in at least one. The first front represents the Pareto optimal set, consisting of individuals that are not dominated by any others. After sorting, the crowding distance for each individual is calculated to ensure diversity. The crowding distance measures how close a solution is to its neighbors in the objective space, based on Euclidean distance in each objective dimension. This encourages maintaining diversity by favoring solutions in less crowded regions.

During selection, individuals are chosen from the sorted fronts in order, starting with the first, until the population size limit is reached. If the last front exceeds the limit, individuals from that front are selected based on their crowding distances. This ensures a well-distributed and diverse population, helping the algorithm converge effectively towards the Pareto optimal front.

\subsection{Summary of Proposed Algorithm}
The AHMOA algorithm effectively integrates memory-based evaluation, adaptive hybrid strategies, and robust objective evaluation to address the challenges of urban traffic management. By considering multiple objectives and incorporating robustness into the optimization process, the proposed method provides a reliable and efficient solution for optimizing traffic signal timings in dynamic urban environments. This comprehensive approach allows our algorithm to effectively navigate the complex, multi-objective landscape of traffic signal optimization while maintaining robustness in the face of dynamic and uncertain traffic conditions. The detail pseudo-codes are given in Algorithm \ref{alg:AHMOA}.

\begin{algorithm}[!htp]
	\caption{AHMOA Traffic Optimization Algorithm}
	\label{alg:AHMOA}
	\begin{algorithmic}[1]
		\STATE \textbf{Input:}
		\begin{itemize}
			\item \textbf{C:} Cycle lengths for each intersection.
			\item \textbf{SR:} Saturation flow rates for each intersection.
			\item \textbf{V:} Traffic volumes for each intersection.
			\item \textbf{AM:} Adjacency matrix indicating connectivity between intersections.
			\item \textbf{IT:} Types of intersections.
		\end{itemize}
		\STATE Initialize population $P$ with random red light ratios $\lambda_i$ for each intersection; Initialize strategy probabilities $\mathbf{p}$ for GA, DE, PSO, and LS
		\FOR{each generation $gen$ from 1 to $maxGen$}
		\FOR{each individual $i$ in population $P$}
		\STATE Evaluate objectives (Average Delay, Network Stability, Robustness) for individual $i$ using historical data $M$
		\ENDFOR
		\STATE Generate offspring using adaptive hybrid strategy by Algorithm \ref{alg:offspring_generation}.
		\STATE Combine parent and offspring populations
		\STATE Perform non-dominated sorting to form Pareto fronts
		\STATE Calculate crowding distances for each individual
		\STATE Select individuals based on fronts and crowding distances until population size is restored
		\STATE Update strategy success rates $\mathbf{R}_{\text{succ}}$ and strategy probabilities $\mathbf{p}$
		\ENDFOR
		\STATE \textbf{Return:}
		\begin{itemize}
			\item \textbf{$\lambda_i$:} Red light ratios of all intersections in each Pareto solution.
			\item \textbf{PS:} Pareto solutions' objectives of average delay, network stability, and robustness.
		\end{itemize}
	\end{algorithmic}
\end{algorithm}

\subsection{Computational Complexity and Scalability}
The computational complexity of the algorithm consists of two main parts: evaluating the objective functions and generating/selecting offspring.

For each generation, the complexity of evaluating the objective functions depends on the number of intersections ($N$) and time steps ($T$), resulting in $O(NT)$. With $E$ evaluations per solution for robustness, the total complexity becomes $O(ENT)$ per generation. For offspring generation, AHMOA uses an adaptive hybrid strategy, dynamically applying GA, DE, PSO, or LS. The complexity per individual is $O(N)$ for GA, DE, and PSO, while LS may increase it to $O(N^2)$. Thus, for a population size $P$, the offspring generation complexity is about $O(PN)$. The selection process uses non-dominated sorting, which has a complexity of $O(P^2M)$, where $M$ is the number of objectives, plus $O(P \log P)$ for crowding distance calculation. The total complexity per generation is $O(PN + P^2M + P \log P)$.

AHMOA scales well for large urban traffic networks due to its adaptive nature. As $N$ and $P$ grow, the algorithm dynamically adjusts strategy probabilities, efficiently balancing exploration and exploitation. Memory-based evaluations further enhance robustness by leveraging historical data without significant computational overhead. Therefore, in large urban networks with thousands of intersections and dynamic traffic patterns, AHMOA remains a scalable solution, capable of handling high-dimensional optimization challenges and optimizing traffic signal timings effectively.

\section{Experimental Results and Discussion}
\label{sec:exp}
In this section, we introduce traffic simulation models and design models tailored to specific city patterns. Based on these models, we compare the performance of the proposed algorithm with several other evolutionary algorithms. The experimental results are analyzed and explained within the context of various traffic scenarios.

\subsection{Traffic Simulation Model}
In the experiments, we employ an intersection-based model to simulate traffic flow dynamics. The Traffic Data Generator simulates traffic conditions for thousands of intersections over a 24-hour period. Each intersection is assigned a random cycle time from a predefined set of cycle times: {60, 90, 120} seconds. This reflects the traffic light cycle duration at each intersection. A base saturation flow rate, representing the maximum number of vehicles per hour that can pass through an intersection when the traffic light is green, is set between 800 and 2400 vehicles per hour for each intersection, varying by intersection type, collector and arterial respectively. During morning peak hours and evening peak hours, the flow is increased by a factor of 1.5, adjusted by a random fluctuation between 0.9 and 1.3 times the base saturation flow. During off-peak hours, the flow is a random value between 50\% and 100\% of the base saturation flow, reflecting typical traffic variability during less busy times. During lunch hours, the flow is increased by a factor of 1.1, adjusted by a random fluctuation between 0.8 and 1.2 times the base saturation flow. During midnight hours, the flow is decreased to 50\% of the base saturation flow.

\subsection{City-Specific Adaptations}
To validate the effectiveness of our proposed algorithm across diverse urban environments, we recognize that each city has distinct traffic patterns and infrastructure characteristics. Our adaptation strategy considers network topology, intersection-based model parameters, dynamic traffic demand, environmental factors, and unique intersection characteristics. By incorporating these city-specific adaptations, our model provides more realistic and effective traffic management solutions across diverse urban environments. Four cities are inovled in the simulations, which are Manhanttan, Istanbul, Paris and São Paulo. These differences highlight the unique challenges and strategies required in traffic planning and management for each city, underscoring the importance of tailored approaches in urban traffic optimization. The features of the four cities are summarized in Table \ref{tab:city_characteristics}. The related figures are presented in Fig. \ref{fig:urban_networks}. 

Given the large-scale traffic network in this paper, a three-dimensional force-directed layout is used for visualization, which ensures a uniform layout of intersection nodes, making it easier to visualize connections and identify clusters of closely connected nodes. In the figures, nodes represent intersections, with colors indicating road types: green for local roads, orange for collector roads, and red for arterial roads. Node size reflects average traffic flow, with larger nodes indicating higher volumes. Blue edges represent connections between intersections, with edge thickness corresponding to average saturation flow, and thicker lines indicate higher saturation.

\begin{table*}[htbp]
	\centering
	\caption{City-Specific Simulation Characteristics}
	\begin{tabular}{cccc}
		\toprule
		\textbf{City} & \textbf{Network Topology} & \textbf{Peak Hours} & \textbf{Traffic Patterns} \\
		\midrule
		Manhattan & Grid system with avenues and streets & 7-9 AM, 5-7 PM & High pedestrian activity, lunchtime peak  \\
		\hline
		Istanbul & Complex network with bridges and tunnels & 7-10 AM, 4-8 PM & Extended peaks, tourist season impact \\
		\hline
		Paris & Radial-concentric with ring roads & 7-9 AM, 5-7 PM & Multiple daily peaks, tourist activity \\
		\hline
		São Paulo & Mix of wide avenues and local streets & 6-10 AM, 4-9 PM & Extended peaks, weekend variations  \\
		\bottomrule
	\end{tabular}
	\label{tab:city_characteristics}
\end{table*}

\subsubsection{Manhattan}
Manhattan's traffic network, presented in Fig. \ref{fig:manhattan}, is characterized by a rigid grid system with 12 north-south avenues and 220 east-west streets. The roads are categorized into local, collector, and arterial types, with major arterials like 5th Avenue, 8th Avenue, 14th Street, and 23rd Street playing crucial roles in traffic flow. This regular and symmetrical grid-like structure reflects the city's typical square block design. Manhattan's traffic patterns include distinct morning (7-9 AM) and evening (5-7 PM) peaks, with a notable increase in activity during lunchtime. The city also experiences high pedestrian activity, which influences traffic flow. The grid structure offers several advantages, such as facilitating even distribution of traffic flow, reducing congestion, and simplifying traffic signal control and optimization. However, during peak hours, intersections can become bottlenecks, leading to congestion at multiple points.

\subsubsection{Istanbul}
Istanbul's traffic network, given in Fig.~\ref{fig:sao_paulo}, is uniquely positioned, spanning two continents, and is characterized by a mix of ancient, narrow streets and modern, wide boulevards. The simulation model includes 40 major roads, categorized into local, collector, and arterial types, incorporating key bridges such as the Bosphorus Bridge and Fatih Sultan Mehmet Bridge, along with numerous secondary roads. The network shows a complex and irregular pattern with multiple centers, indicative of the city's historical street layout and varied terrain. Istanbul experiences significant congestion during both morning (7-10 AM) and evening (4-8 PM) peak hours. The city's hilly terrain and seasonal traffic fluctuations are incorporated into the model. The multi-centered traffic network of Istanbul helps in dispersing traffic flow, potentially reducing congestion in specific areas. However, the irregular network increases the complexity of traffic management, making it challenging to implement uniform traffic signal control optimization.

\subsubsection{Paris}
Paris features a combination of wide boulevards and narrow streets, with a series of ring roads (périphériques). The figure is presented in Fig. \ref{fig:paris}. The city's traffic network is simulated with 20 major roads, categorized into local, collector, and arterial types, incorporating ring road connections. Paris exhibits a distinctive ring-radial structure, expanding from the center outwards, mirroring the city's radial avenues extending from central points like the Arc de Triomphe. The city experiences multiple traffic peaks throughout the day, with primary peak hours from 7-9 AM and 5-7 PM. Significant tourist activity in central areas also impacts traffic patterns, with reduced activity on weekends. The radial structure of Paris aids in distributing traffic across multiple directions, preventing congestion in any single direction, and facilitates both centralized and decentralized traffic management. However, during peak periods, this structure can lead to bottlenecks in central areas, particularly at key junctions where the ring and radial lines intersect.

\subsubsection{São Paulo}
\begin{figure}[!htbp]
	\centering
	\begin{subfigure}[b]{0.45\linewidth}
		\centering
		\includegraphics[width=\textwidth, trim={35mm 35mm 35mm 35mm}, clip]{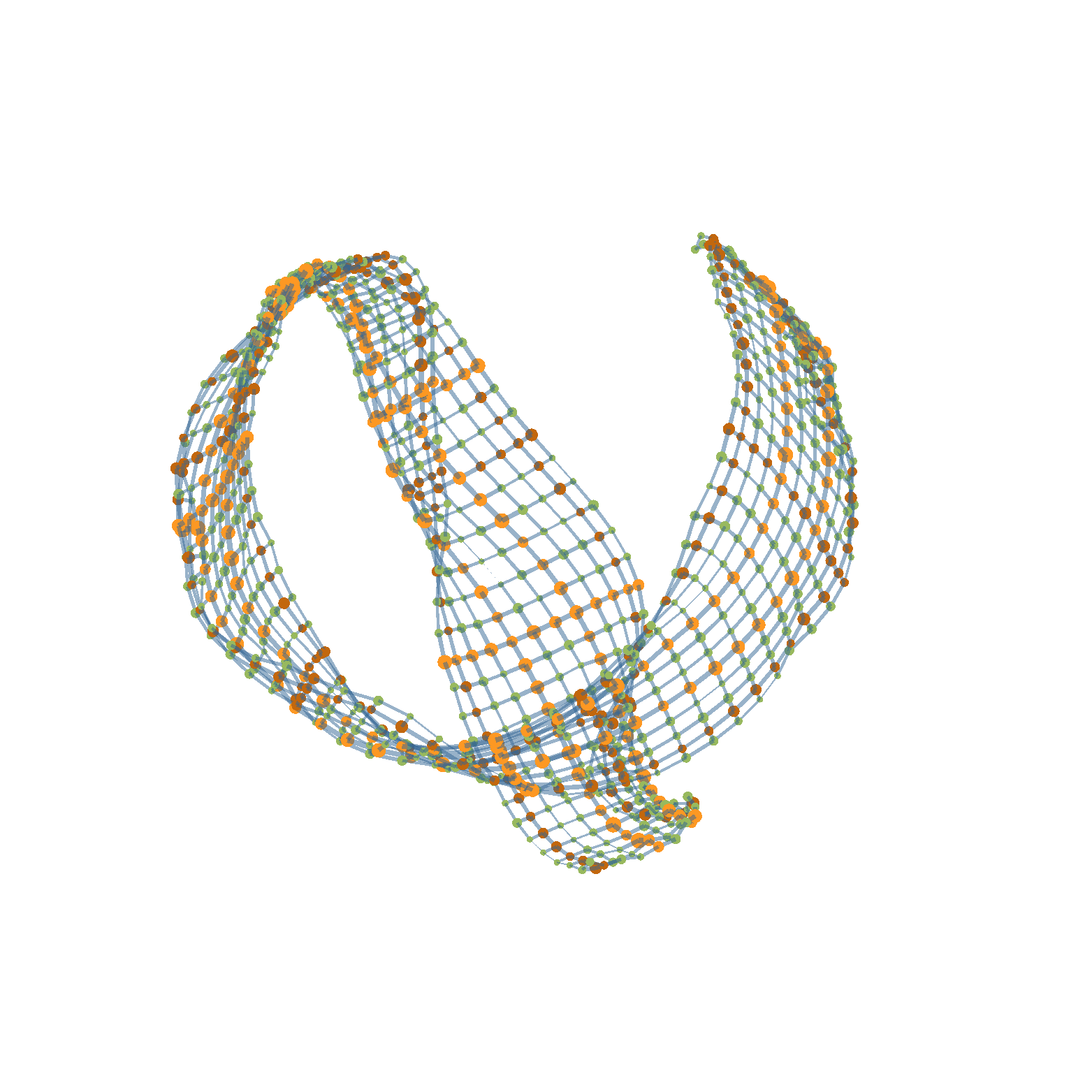}
		\caption{Manhattan Traffic Network}
		\label{fig:manhattan}
	\end{subfigure}
	\hfill
	\begin{subfigure}[b]{0.45\linewidth}
		\centering
		\includegraphics[width=\textwidth, trim={35mm 35mm 35mm 35mm}, clip]{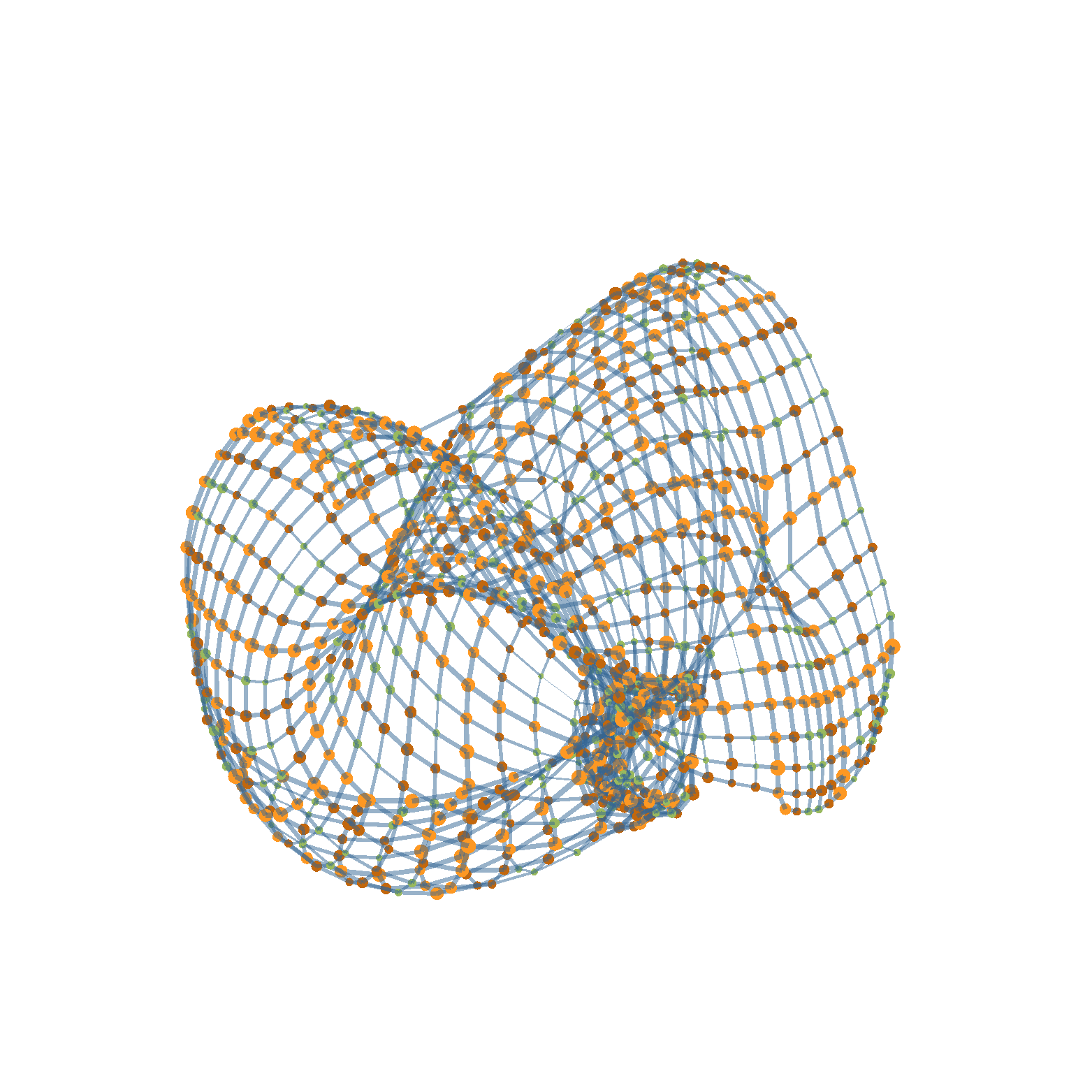}
		\caption{Istanbul Traffic  Network}
		\label{fig:istanbul}
	\end{subfigure}
	\vfill
	\begin{subfigure}[b]{0.45\linewidth}
		\centering
		\includegraphics[width=\textwidth, trim={35mm 35mm 35mm 35mm}, clip]{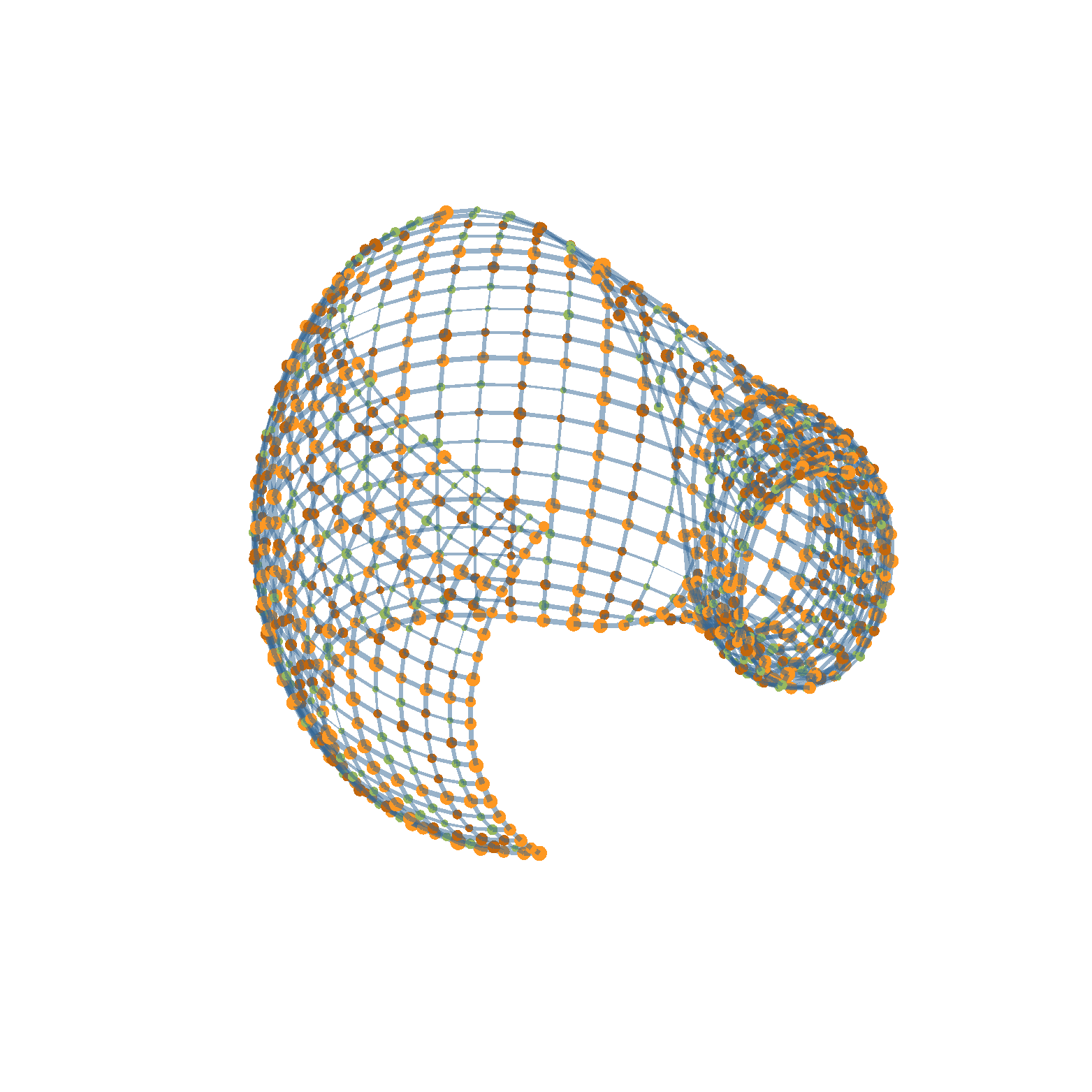}
		\caption{Paris Traffic Network}
		\label{fig:paris}
	\end{subfigure}
	\hfill
	\begin{subfigure}[b]{0.45\linewidth}
		\centering
		\includegraphics[width=\textwidth, trim={35mm 35mm 35mm 35mm}, clip]{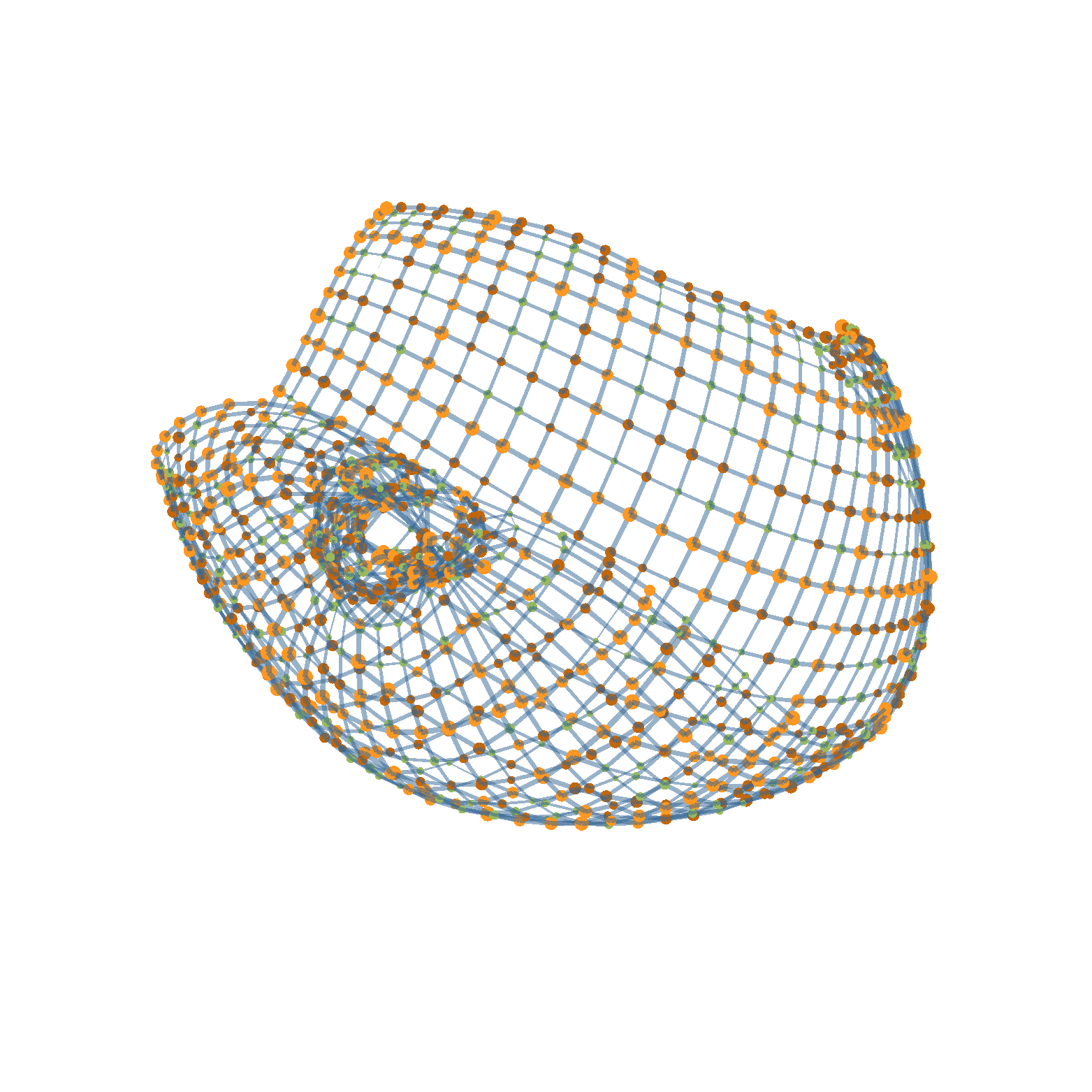}
		\caption{São Paulo Traffic Network}
		\label{fig:sao_paulo}
	\end{subfigure}
	\caption{Comparison of different urban networks}
	\label{fig:urban_networks}
\end{figure}

São Paulo's traffic network, shown in Fig. \ref{fig:sao_paulo}, is characterized by a less uniform street layout, featuring a mix of wide avenues and local streets. The simulation model includes 30 major roads and 50 secondary roads, categorized into local, collector, and arterial types, including high-capacity road connections. The city's traffic network presents a complex and irregular mesh structure, influenced by diverse terrain and uneven urban development. São Paulo experiences severe congestion, especially during extended peak hours (6-10 AM and 4-9 PM). Traffic patterns account for extended peak hours, weekend variations, and the impact of seasonal weather. The irregular mesh network of São Paulo can adapt flexibly to terrain and urban growth, dispersing traffic flow and potentially reducing congestion in specific areas. However, the complexity of the network increases the difficulty of traffic management and optimization, potentially leading to local bottlenecks and congestion if not systematically planned.

\subsection{Experimental Results}
In the experiments, the AHMOA algorithm is designed to optimize large-scale urban traffic systems with multiple objectives. The population size is set to 120, and the algorithm runs for a maximum of 50 generations. It optimizes three objectives, evaluating each solution five times to account for uncertainties such as weather and traffic conditions, with simulations covering a full 24-hour period. In the experimental comparison, we used three peer algorithms: MOEA/D \cite{zhang2007moead}, NSGA3 \cite{deb2014nsga3}, and NSDE3. In NSDE3, the genetic algorithm in NSGA3 is replaced with a differential evolution algorithm. The results are presented and compared as follows. 

\subsubsection{Experiments on Manhattan Traffic Network}
For the Manhattan scenario, we aggregate the solutions obtained by AHMOA, MOEA/D, NSGA3, and NSDE3, and through dominating rules, identified the global Pareto optimal solutions, which are presented in Table~\ref{tab:glo_manhattan}. The data in the table have been processed in logarithmic form to facilitate comparison of the values. As shown in the table, all 17 global Pareto optimal solutions are obtained by the AHMOA algorithm, further demonstrating the superiority of the proposed method.
\begin{table}[htbp]
	\centering
	\caption{Global Pareto Solutions for Manhattan}
	\begin{adjustbox}{max width=\textwidth}
		\begin{tabular}{cccc}
			\toprule
			\textbf{Algorithm} & \textbf{Log(Objective 1)} & \textbf{Log(Objective 2)} & \textbf{Log(Objective 3)} \\
			\midrule
			\multirow{17}{*}{AHMOA}
			& 8.0416 & 7.9853 & -8.1186 \\
			& 8.0418 & 7.9851 & -7.8175 \\
			& 8.0418 & 7.9852 & -Inf \\
			& 8.0419 & 7.9848 & -7.8175 \\
			& 8.0420 & 7.9849 & -8.1186 \\
			& 8.0421 & 7.9846 & -7.5165 \\
			& 8.0421 & 7.9850 & -Inf \\
			& 8.0422 & 7.9846 & -8.1186 \\
			& 8.0424 & 7.9843 & -7.6414 \\
			& 8.0424 & 7.9844 & -8.1186 \\
			& 8.0427 & 7.9841 & -7.8175 \\
			& 8.0427 & 7.9843 & -8.1186 \\
			& 8.0429 & 7.9842 & -8.1186 \\
			& 8.0431 & 7.9840 & -8.1186 \\
			& 8.0435 & 7.9841 & -Inf \\
			& 8.0438 & 7.9839 & -7.6414 \\
			& 8.0440 & 7.9840 & -Inf \\
			\bottomrule
		\end{tabular}
	\end{adjustbox}
	\label{tab:glo_manhattan}
\end{table}

\begin{figure*}[!htbp]
	\centering
	\begin{subfigure}[!htbp]{0.16\linewidth}
		\centering
		\includegraphics[width=\textwidth, trim={270mm 10mm 270mm 10mm}, clip]{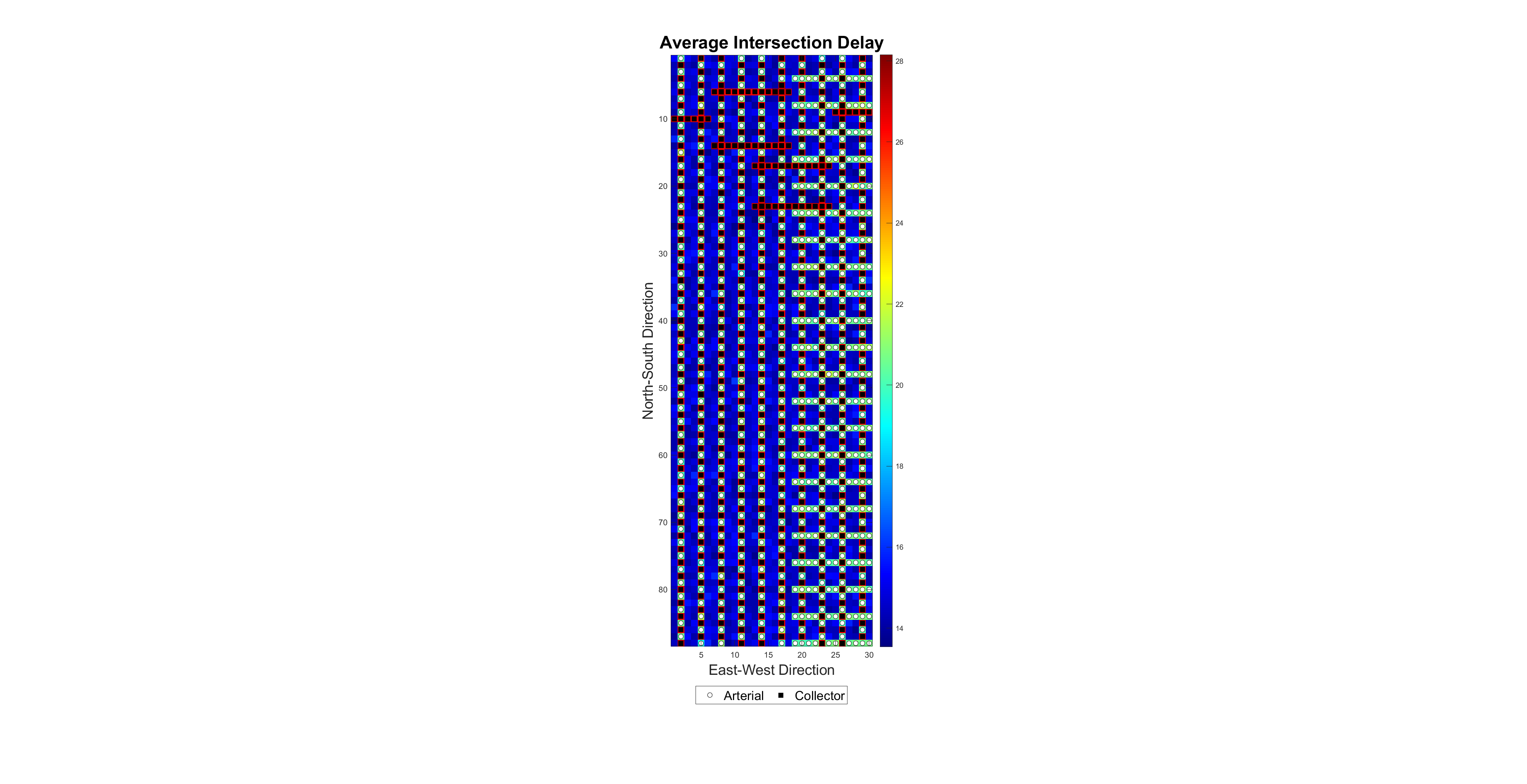}
		\caption{Baseline}
		\label{fig:m-before}
	\end{subfigure}
	\hfill
	\begin{subfigure}[!htbp]{0.16\linewidth}
		\centering
		\includegraphics[width=\textwidth, trim={270mm 10mm 270mm 10mm}, clip]{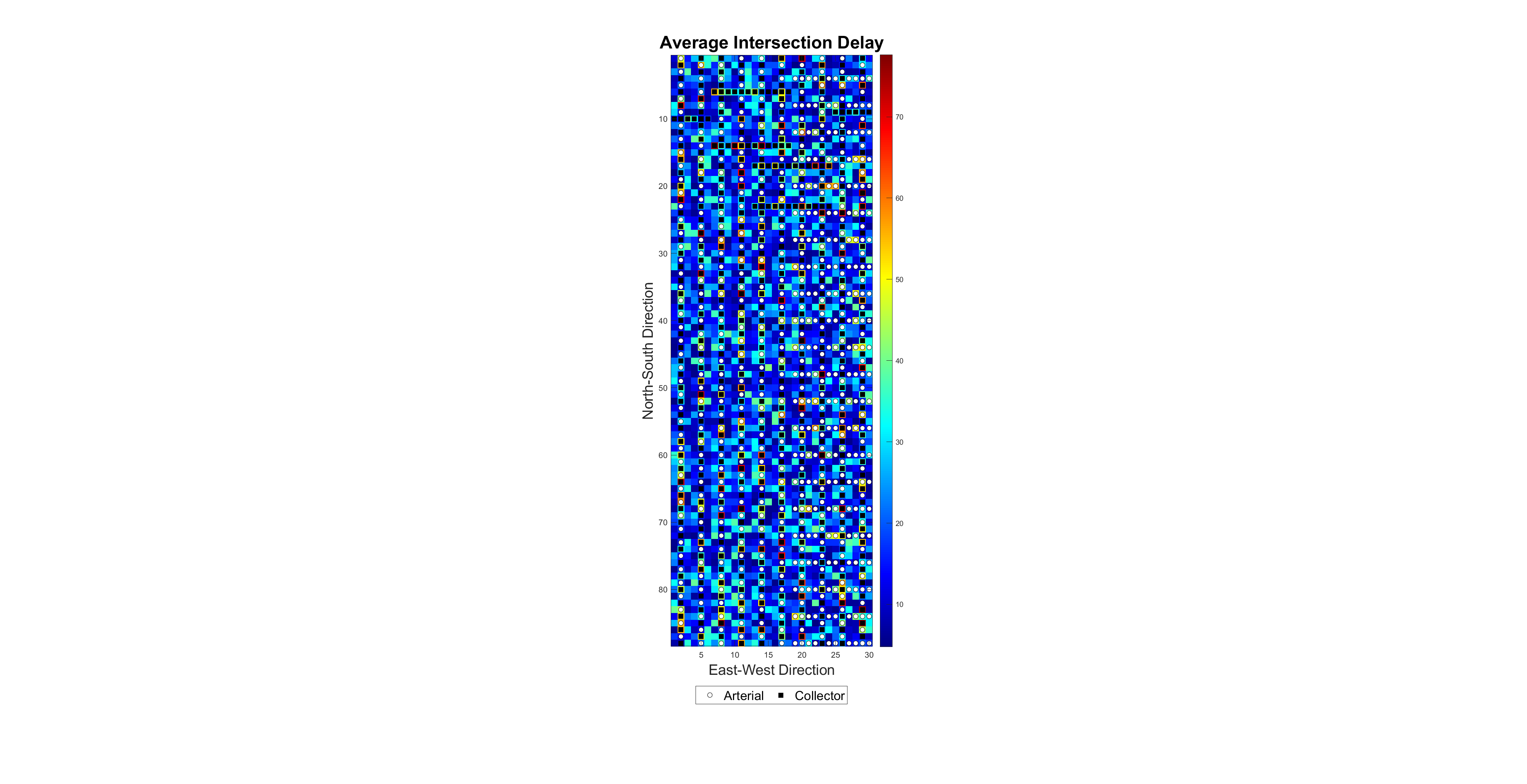}
		\caption{NSGA-III}
		\label{fig:m-nsga}
	\end{subfigure}
	\hfill
	\begin{subfigure}[!htbp]{0.16\linewidth}
		\centering
		\includegraphics[width=\textwidth, trim={270mm 10mm 270mm 10mm}, clip]{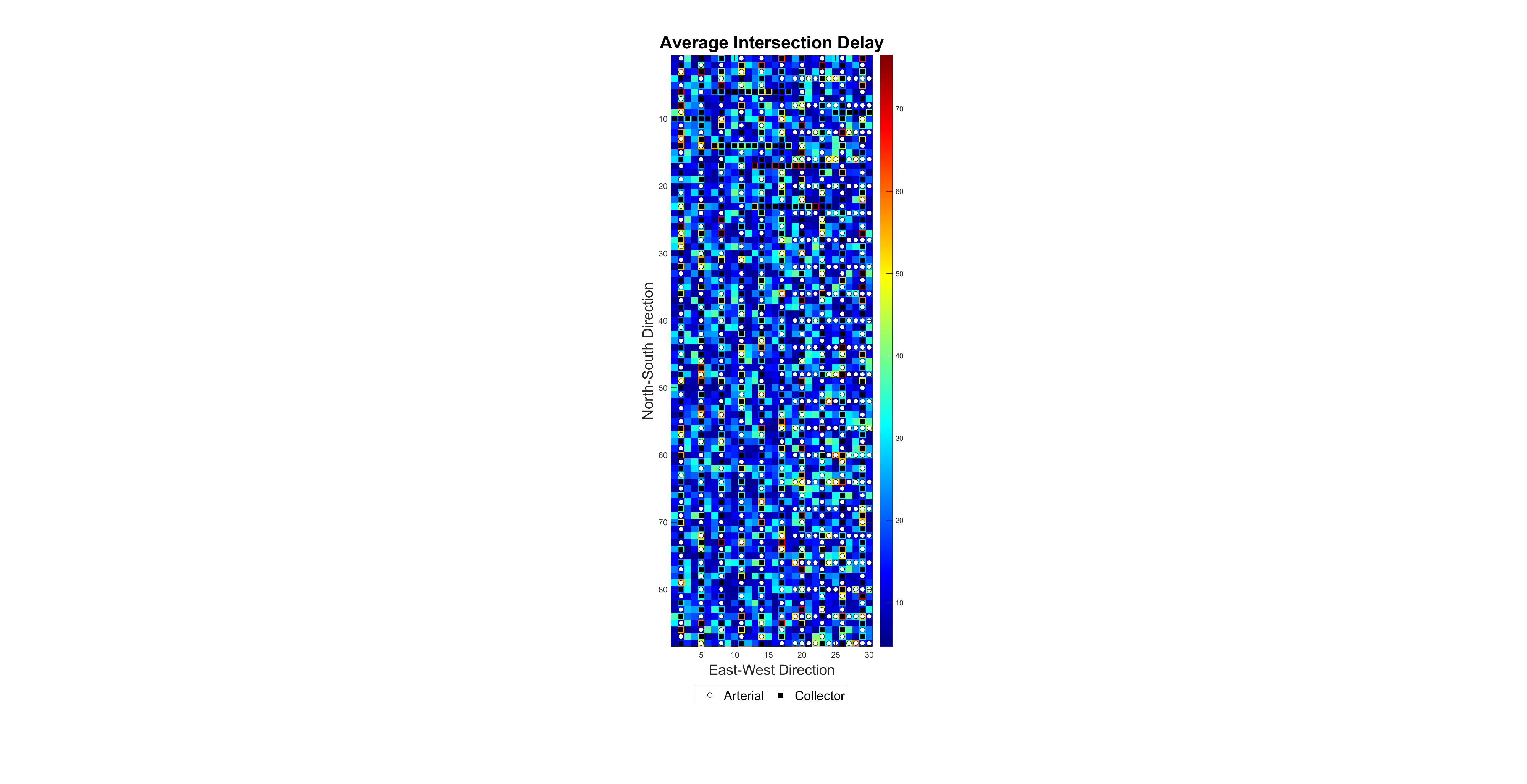}
		\caption{NSDE}
		\label{fig:m-nsde}
	\end{subfigure}
	\hfill
	\begin{subfigure}[!htbp]{0.16\linewidth}
		\centering
		\includegraphics[width=\textwidth, trim={270mm 10mm 270mm 10mm}, clip]{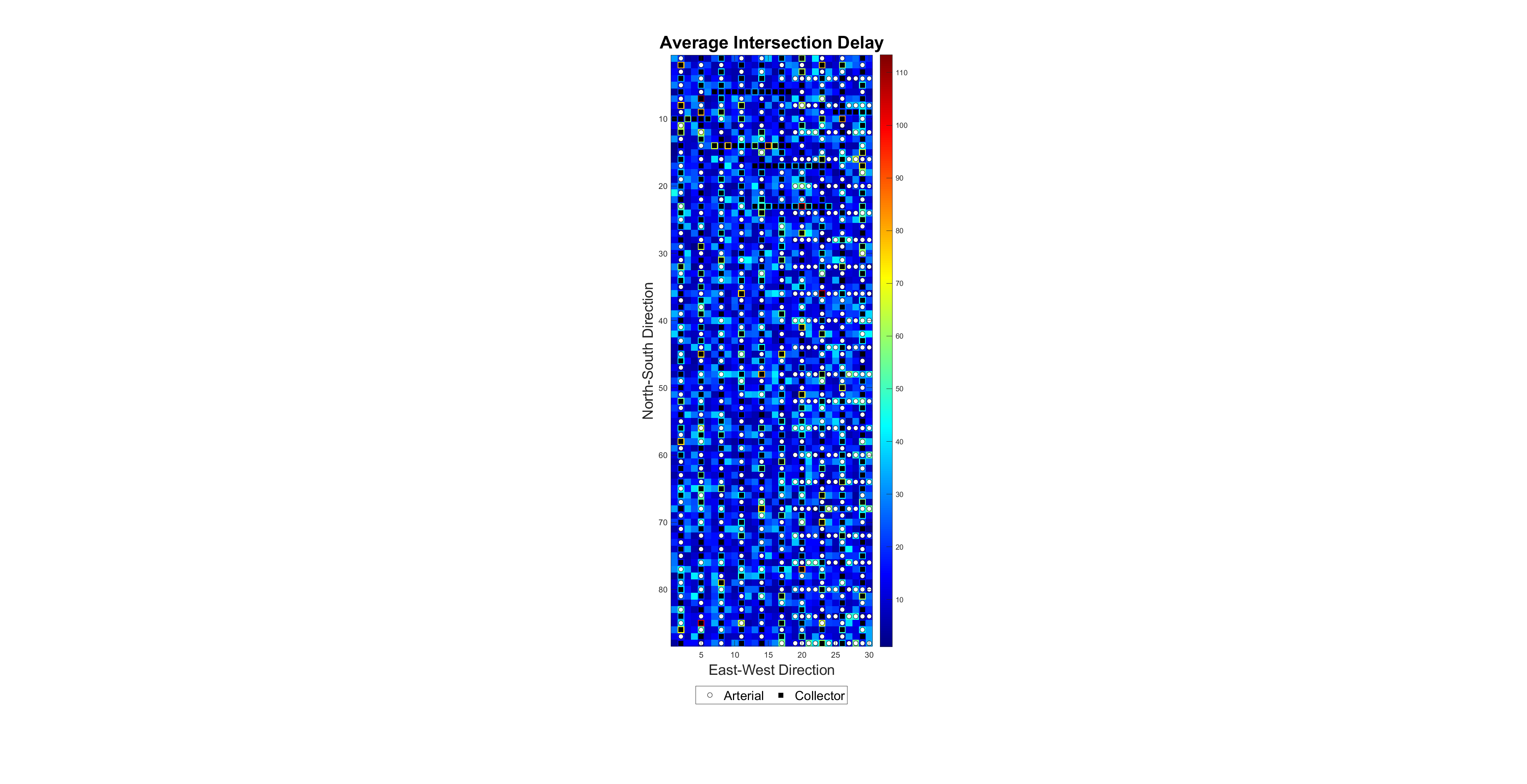}
		\caption{MOEA/D}
		\label{fig:m-moead}
	\end{subfigure}
	\hfill
	\begin{subfigure}[!htbp]{0.16\linewidth}
		\centering
		\includegraphics[width=\textwidth, trim={270mm 10mm 270mm 10mm}, clip]{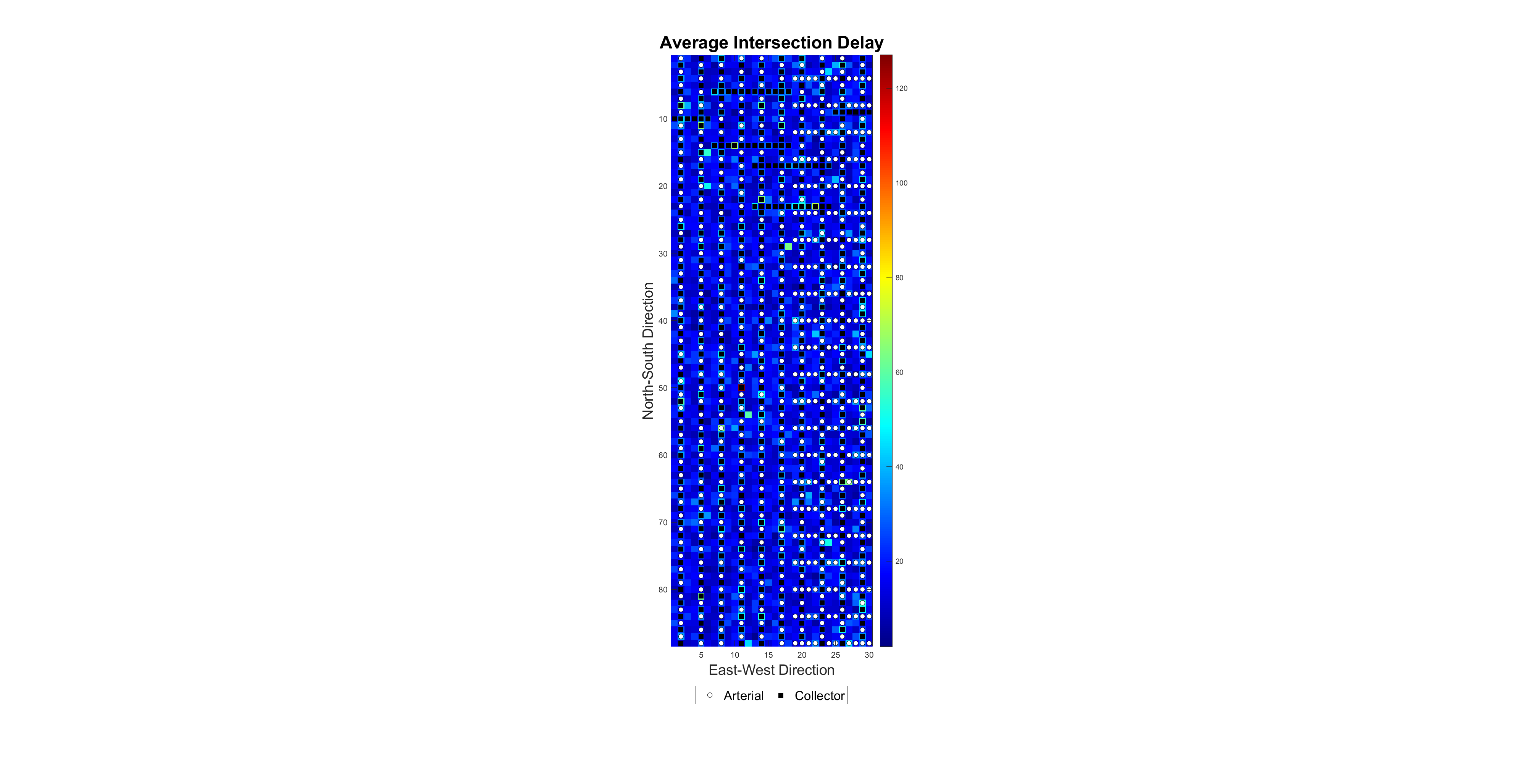}
		\caption{AHMOA}
		\label{fig:m-ahmoa}
	\end{subfigure}
	\caption{Comparison of Algorithm Performances on Average Intersection Delay in Manhattan City}
	\label{fig:intersectiondelay_m}
\end{figure*}

To provide a clearer performance comparison, we present Fig.~\ref{fig:intersectiondelay_m}, where we have selected a solution from the Pareto front of each algorithm for scenario-based visualization, reflecting the average delay during 24 hours. Fig.~\ref{fig:m-before} illustrates the average delay at each intersection over a 24-hour period before optimization. In the optimization simulation, we designed 22 arterial roads (marked as small circles) and 120 collectors (marked as small black squares), resulting in a total of 2,640 intersections. However, for ease of visualization and layout, the figures display these intersections in a grid of size 88x30. In the figure, the color gradient from red to blue represents delay values ranging from high to low. In the Baseline scenario, there is a significant presence of red areas, indicating high average delays in the Manhattan area without optimization. The performance of the AHMOA algorithm is shown in Fig.~\ref{fig:m-ahmoa}, where it is evident that the entire region is predominantly blue, reflecting minimal delays and demonstrating the effectiveness of the AHMOA algorithm. The performance of MOEA/D is displayed in Fig.~\ref{fig:m-moead}, where the region is also mostly blue, but there are more scattered orange and green areas compared to AHMOA, indicating relatively higher delays in some locations. Finally, Fig.~\ref{fig:m-nsga} and Fig.~\ref{fig:m-nsde} show the performance of the NSGA and NSDE algorithms, respectively. While these algorithms achieve some optimization, the presence of more red, orange, and green areas suggests that their performance is still suboptimal compared to AHMOA, with higher delays in several parts of the network.

\subsubsection{Experiments on Istanbul Traffic Network}
In the Istanbul scenario, we set 30 arterial roads and 50 collectors. The peak traffic periods, from 7 AM to 10 AM and 4 PM to 8 PM, increase flow by 20-30\%. The adjacency matrix defines the network's connectivity, including bridges and tunnels. Network visualization highlights traffic intensity through node sizes and saturation flow through edge widths, with arterial, collector, and local roads shown in distinct colors. The global Pareto solutions obtained by the four algorithms are shown in Table \ref{tab:glo_istanbul}. From this table, we observe that the global Pareto solutions are obtained by only two algorithms: MOEA/D and our proposed AHMOA, both yielding a similar number of feasible solutions. By comparing the data, we find that MOEA/D tends to generate solutions with lower delays, while AHMOA provides better solutions in terms of network stability and robustness. On the other hand, NSGA3 and NSDE3 fail to obtain any global Pareto solutions in this scenario.

\begin{table}[htbp]
	\centering
	\caption{Global Pareto Solutions for Istanbul}
	\begin{adjustbox}{max width=\textwidth}
		\begin{tabular}{cccc}
			\toprule
			\textbf{Algorithm} & \textbf{Log(Objective 1)} & \textbf{Log(Objective 2)} & \textbf{Log(Objective 3)} \\
			\midrule
			\multirow{20}{*}{MOEAD} & 8.1696 & 7.8863 & -7.8175 \\
			& 8.1703 & 7.8859 & -7.8175 \\
			& 8.1711 & 7.8857 & -7.4196 \\
			& 8.1712 & 7.8858 & -7.5165 \\
			& 8.1714 & 7.8855 & -7.8175 \\
			& 8.1718 & 7.8857 & -8.1186 \\
			& 8.1720 & 7.8846 & -7.6414 \\
			& 8.1727 & 7.8847 & -7.8175 \\
			& 8.1728 & 7.8843 & -7.8175 \\
			& 8.1729 & 7.8850 & -\text{Inf} \\
			& 8.1732 & 7.8842 & -7.4196 \\
			& 8.1738 & 7.8842 & -7.5165 \\
			& 8.1740 & 7.8841 & -7.3404 \\
			& 8.1741 & 7.8835 & -7.6414 \\
			& 8.1747 & 7.8835 & -7.8175 \\
			& 8.1750 & 7.8828 & -7.5165 \\
			& 8.1757 & 7.8830 & -7.6414 \\
			& 8.1801 & 7.8824 & -7.1643 \\
			& 8.1802 & 7.8825 & -7.2155 \\
			& 8.1803 & 7.8824 & -7.3404 \\
			\midrule
			\multirow{21}{*}{AHMOA} & 8.1813 & 7.8642 & -7.6414 \\
			& 8.1818 & 7.8635 & -7.4196 \\
			& 8.1820 & 7.8636 & -8.1186 \\
			& 8.1822 & 7.8633 & -7.6414 \\
			& 8.1822 & 7.8634 & -8.1186 \\
			& 8.1825 & 7.8627 & -7.4196 \\
			& 8.1825 & 7.8635 & -\text{Inf} \\
			& 8.1826 & 7.8631 & -\text{Inf} \\
			& 8.1828 & 7.8625 & -7.4196 \\
			& 8.1829 & 7.8628 & -7.5165 \\
			& 8.1830 & 7.8626 & -7.8175 \\
			& 8.1830 & 7.8630 & -8.1186 \\
			& 8.1831 & 7.8630 & -\text{Inf} \\
			& 8.1833 & 7.8624 & -7.8175 \\
			& 8.1833 & 7.8628 & -\text{Inf} \\
			& 8.1835 & 7.8625 & -8.1186 \\
			& 8.1836 & 7.8623 & -7.3404 \\
			& 8.1837 & 7.8623 & -7.6414 \\
			& 8.1837 & 7.8624 & -\text{Inf} \\
			& 8.1839 & 7.8622 & -7.4196 \\
			& 8.1841 & 7.8621 & -7.5165 \\
			& 8.1841 & 7.8622 & -\text{Inf} \\
			\bottomrule
		\end{tabular}
	\end{adjustbox}
	\label{tab:glo_istanbul}
\end{table}

\begin{figure*}[!htbp]
	\centering
	\begin{subfigure}[!htbp]{0.16\linewidth}
		\centering
		\includegraphics[width=\textwidth, trim={0mm 0mm 0mm 0mm}, clip]{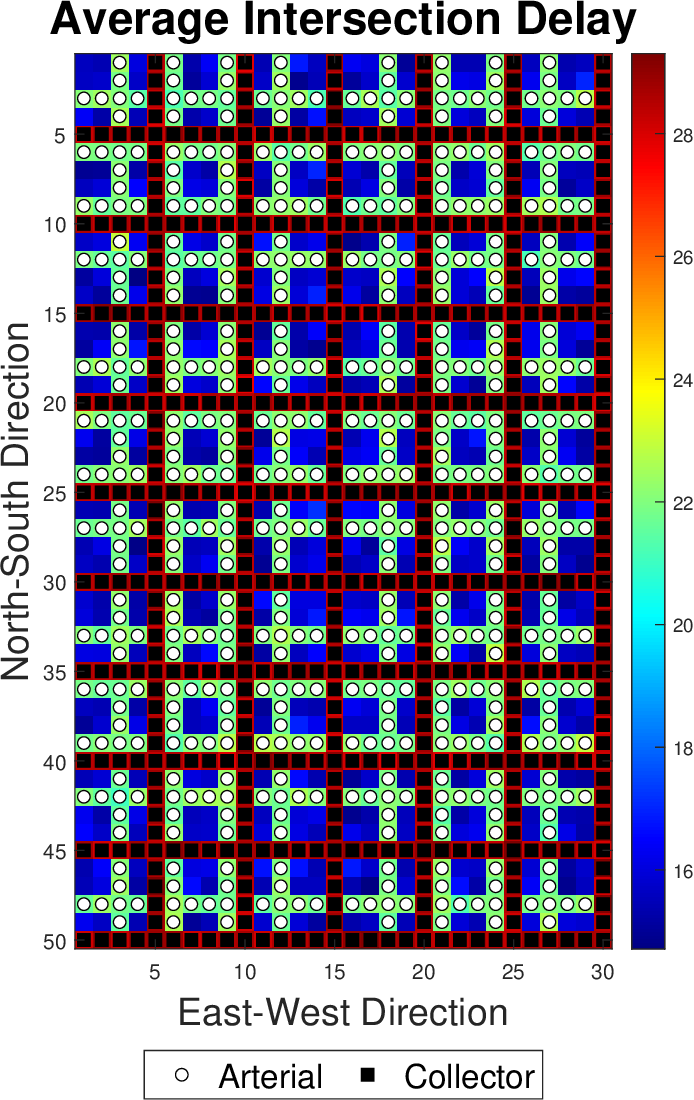}
		\caption{Baseline}
		\label{fig:i-before}
	\end{subfigure}
	\hfill
	\begin{subfigure}[!htbp]{0.16\linewidth}
		\centering
		\includegraphics[width=\textwidth, trim={0mm 0mm 0mm 0mm}, clip]{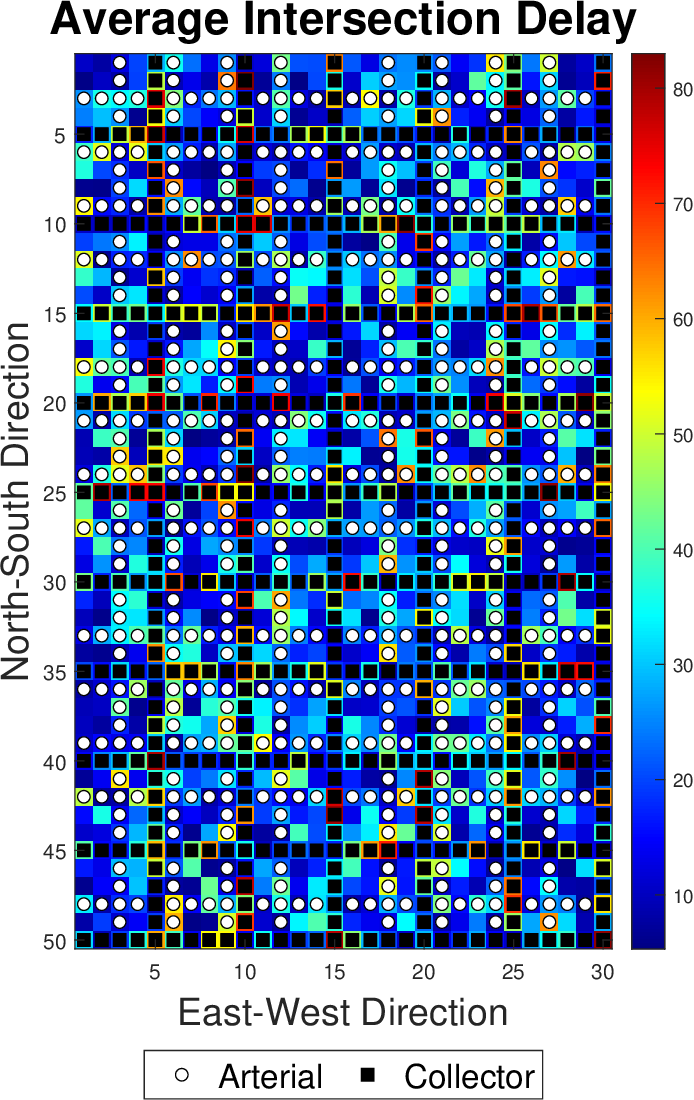}
		\caption{NSGA-III}
		\label{fig:i-nsga}
	\end{subfigure}
	\hfill
	\begin{subfigure}[!htbp]{0.16\linewidth}
		\centering
		\includegraphics[width=\textwidth, trim={0mm 0mm 0mm 0mm}, clip]{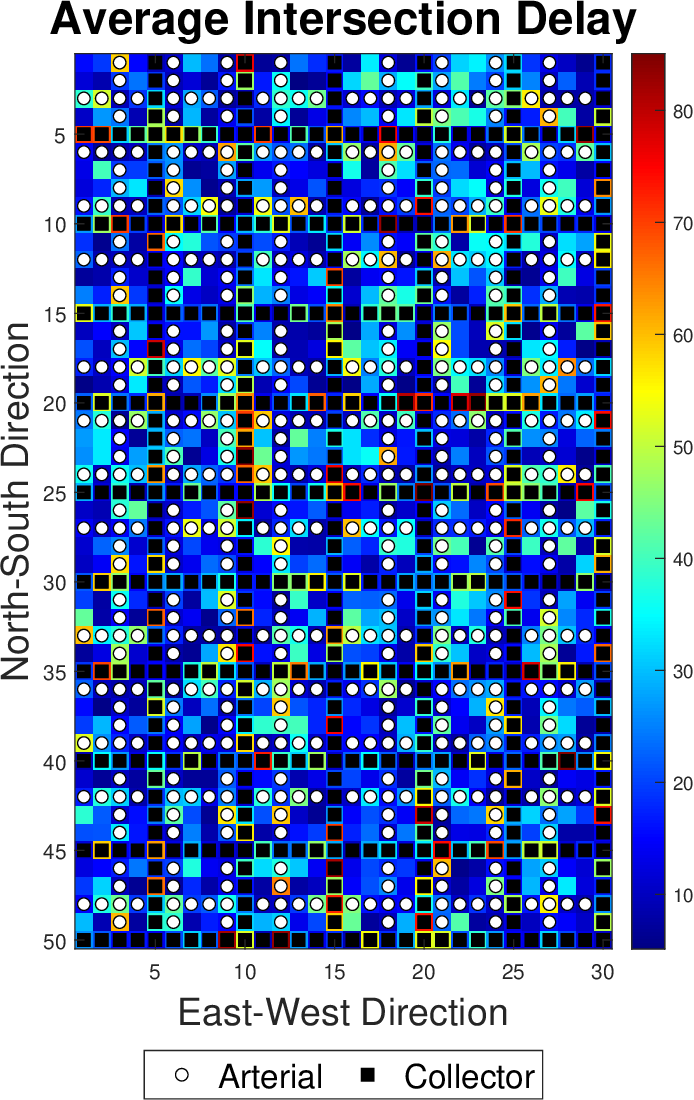}
		\caption{NSDE}
		\label{fig:i-nsde}
	\end{subfigure}
	\hfill
	\begin{subfigure}[!htbp]{0.16\linewidth}
		\centering
		\includegraphics[width=\textwidth, trim={0mm 0mm 0mm 0mm}, clip]{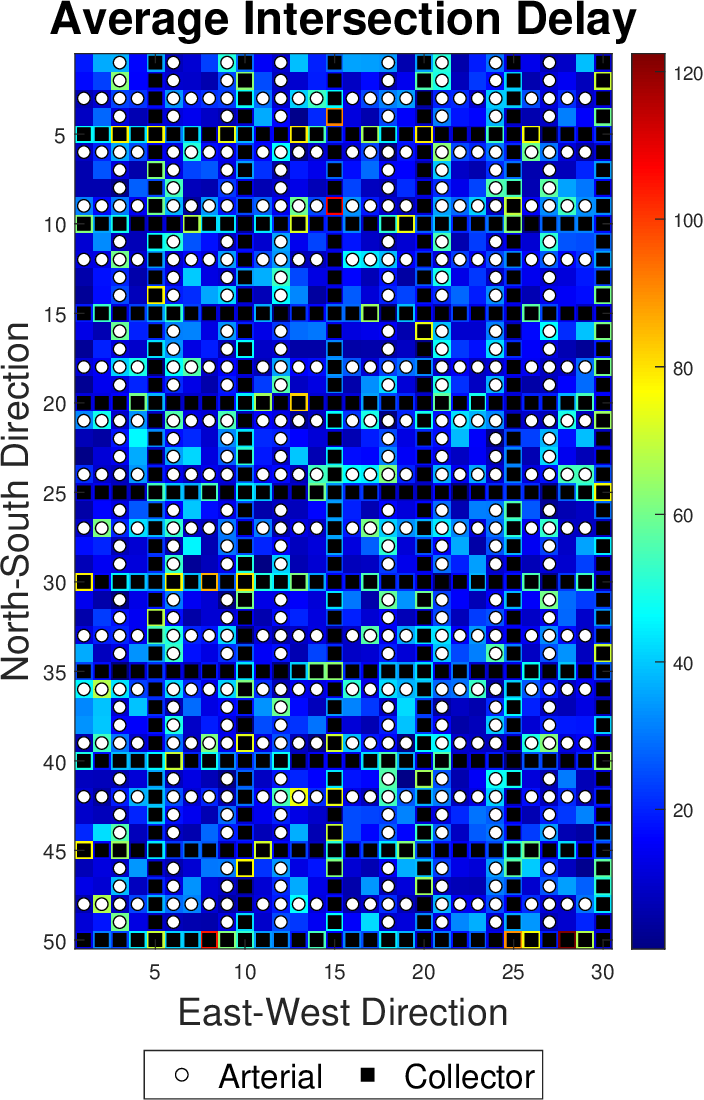}
		\caption{MOEA/D}
		\label{fig:i-moead}
	\end{subfigure}
	\hfill
	\begin{subfigure}[!htbp]{0.16\linewidth}
		\centering
		\includegraphics[width=\textwidth, trim={0mm 0mm 0mm 0mm}, clip]{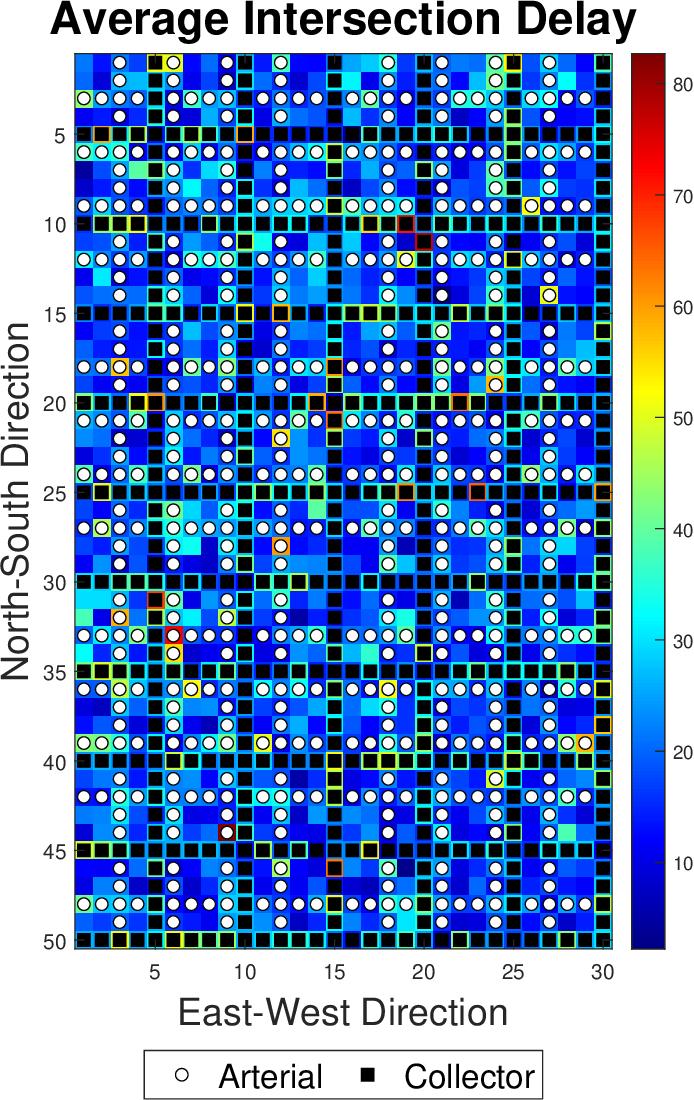}
		\caption{AHMOA}
		\label{fig:i-ahmoa}
	\end{subfigure}
	\caption{Comparison of Algorithm Performances on Average Intersection Delay in Istanbul City}
	\label{fig:intersectiondelay_i}
\end{figure*}

In the performance comparison of average delay, we selected one Pareto solution from each of the four algorithms for visualization. As shown in Fig.~\ref{fig:intersectiondelay_i}, we can observe that the pre-optimization average delay displayed in Fig.~\ref{fig:i-before} contains large red areas indicating high delays. In terms of delay performance, the MOEA/D algorithm achieves slightly more prominent blue regions (shown in Fig.~\ref{fig:i-moead}), while AHMOA exhibits relatively suboptimal delay (shown in Fig.~\ref{fig:i-ahmoa}). This is consistent with the data in Table \ref{tab:glo_istanbul}, indicating that, from the perspective of average delay alone, MOEA/D is superior. However, AHMOA excels in network stability and the robustness of traffic signal control strategies across different scenarios. In NSGA3 and NSDE3 (shown in Fig.~\ref{fig:i-nsga} and Fig.~\ref{fig:i-nsde}), more pronounced green or red areas appear, indicating that average delays remain high in certain regions.

\subsubsection{Experiments on Paris Traffic Network}
The results of the four algorithms are compared to obtain the global Pareto solutions shown in Table \ref{tab:glo_paris}. Due to the large scale of the solutions, the final data in the table are logarithmically processed, keeping only four decimal places after the logarithmic transformation. Identical solutions are removed, and the Pareto fronts are reconstructed to create Table \ref{tab:glo_paris}. The same processing method is applied in the following text. The performance of the four algorithms—MOEA/D, NSGA3, NSDE3, and AHMOA—has been compared to generate the global Pareto solutions, as presented in Table~\ref{tab:glo_paris}. While all algorithms except NSGA3 demonstrate the ability to obtain non-dominated solutions, AHMOA stands out by generating the largest number of global Pareto-optimal solutions. This highlights its strong performance in discovering a broader range of optimal solutions across diverse objectives. 
\begin{table}[htbp]
	\centering
	\caption{Global Pareto Solutions for Paris}
	\begin{adjustbox}{max width=\textwidth}
		\begin{tabular}{cccc}
			\toprule
			\textbf{Algorithm} & \textbf{Log(Objective 1)} & \textbf{Log(Objective 2)} & \textbf{Log(Objective 3)} \\
			\midrule
			\multirow{9}{*}{MOEA/D} 
			& 8.0487 & 7.7730 & -8.4196 \\
			& 8.0498 & 7.7721 & -8.1186 \\
			& 8.0503 & 7.7713 & -8.1186 \\
			& 8.0504 & 7.7718 & -8.4196 \\
			& 8.0505 & 7.7706 & -8.1186 \\
			& 8.0529 & 7.7703 & -8.4196 \\
			& 8.0534 & 7.7696 & -8.1186 \\
			& 8.0619 & 7.7660 & -8.1186 \\
			& 8.0620 & 7.7660 & -8.4196 \\
			\midrule
			\multirow{12}{*}{NSDE3}
			& 8.0460 & 7.7720 & -7.4654 \\
			& 8.0478 & 7.7715 & -7.6414 \\
			& 8.0480 & 7.7684 & -7.9425 \\
			& 8.0512 & 7.7679 & -7.7206 \\
			& 8.0536 & 7.7701 & -8.4196 \\
			& 8.0537 & 7.7681 & -7.8175 \\
			& 8.0542 & 7.7668 & -8.1186 \\
			& 8.0550 & 7.7656 & -7.9425 \\
			& 8.0560 & 7.7691 & -Inf \\
			& 8.0599 & 7.7650 & -7.5745 \\
			& 8.0611 & 7.7645 & -7.8175 \\
			& 8.0613 & 7.7662 & -8.1186 \\
			& 8.0617 & 7.7640 & -7.9425 \\
			\midrule
			\multirow{21}{*}{AHMOA}
			& 8.0652 & 7.7587 & -7.5745 \\
			& 8.0652 & 7.7588 & -7.7206 \\
			& 8.0655 & 7.7553 & -7.8175 \\
			& 8.0658 & 7.7547 & -7.6414 \\
			& 8.0665 & 7.7540 & -7.3782 \\
			& 8.0666 & 7.7543 & -7.5165 \\
			& 8.0670 & 7.7544 & -7.7206 \\
			& 8.0670 & 7.7549 & -8.4196 \\
			& 8.0671 & 7.7540 & -7.8175 \\
			& 8.0674 & 7.7544 & -7.9425 \\
			& 8.0684 & 7.7539 & -7.7206 \\
			& 8.0686 & 7.7533 & -7.9425 \\
			& 8.0688 & 7.7532 & -7.4654 \\
			& 8.0689 & 7.7532 & -7.9425 \\
			& 8.0701 & 7.7531 & -7.4654 \\
			& 8.0704 & 7.7531 & -7.7206 \\
			& 8.0707 & 7.7529 & -7.8175 \\
			& 8.0708 & 7.7531 & -7.9425 \\
			& 8.0712 & 7.7527 & -7.7206 \\
			& 8.0713 & 7.7528 & -7.8175 \\
			& 8.0715 & 7.7528 & -7.9425 \\
			& 8.0723 & 7.7546 & -8.1186 \\
			& 8.0724 & 7.7530 & -Inf \\
			\bottomrule
		\end{tabular}
	\end{adjustbox}
	\label{tab:glo_paris}
\end{table}

\begin{figure*}[!htbp]
	\centering
	\begin{subfigure}[!htbp]{0.17\linewidth}
		\centering
		\includegraphics[width=\textwidth, trim={270mm 10mm 270mm 10mm}, clip]{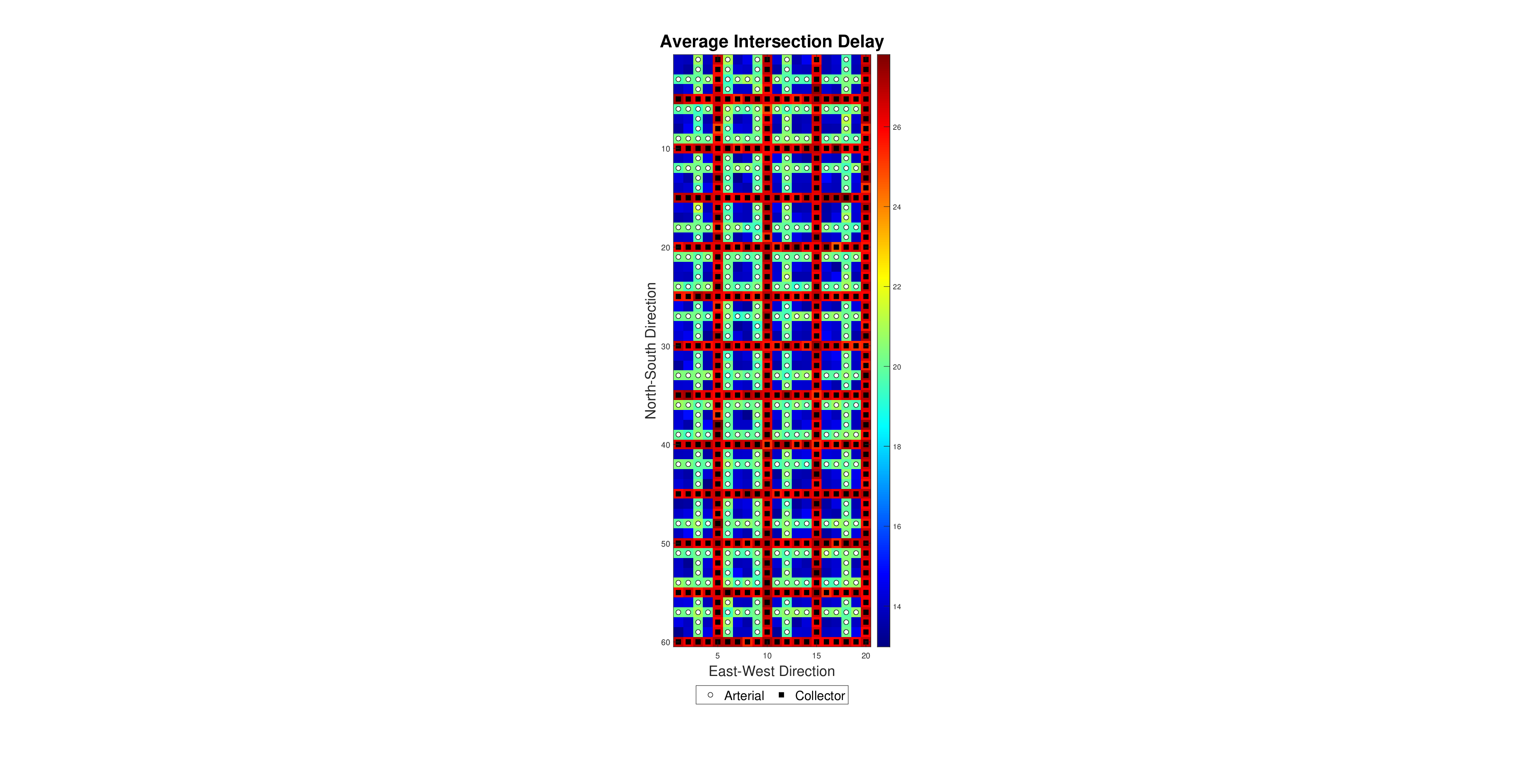}
		\caption{Baseline}
		\label{fig:p-before}
	\end{subfigure}
	\hfill
	\begin{subfigure}[!htbp]{0.17\linewidth}
		\centering
		\includegraphics[width=1\textwidth, trim={270mm 10mm 270mm 10mm}, clip]{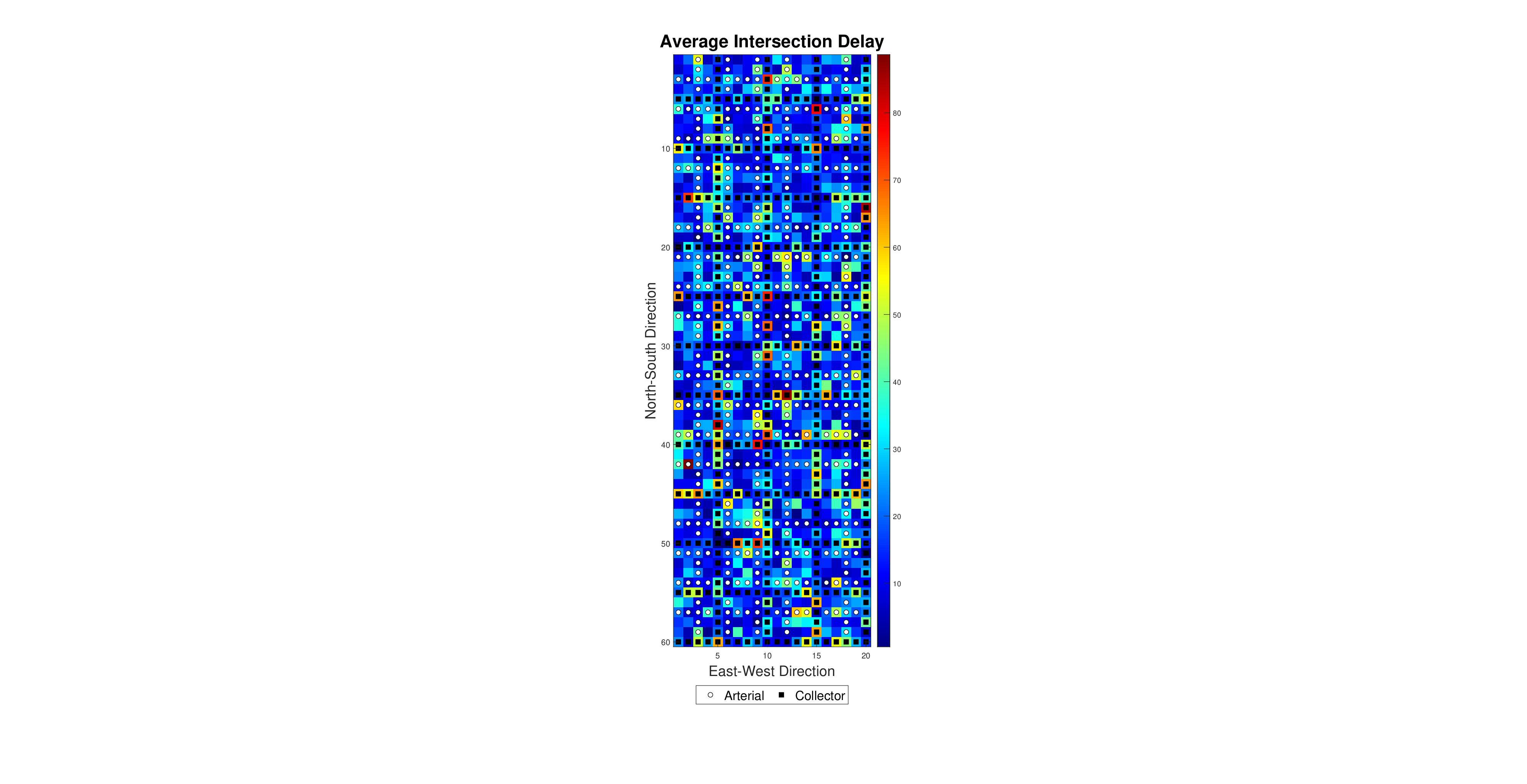}
		\caption{NSGA-III}
		\label{fig:p-nsga}
	\end{subfigure}
	\hfill
	\begin{subfigure}[!htbp]{0.17\linewidth}
		\centering
		\includegraphics[width=1\textwidth, trim={270mm 10mm 270mm 10mm}, clip]{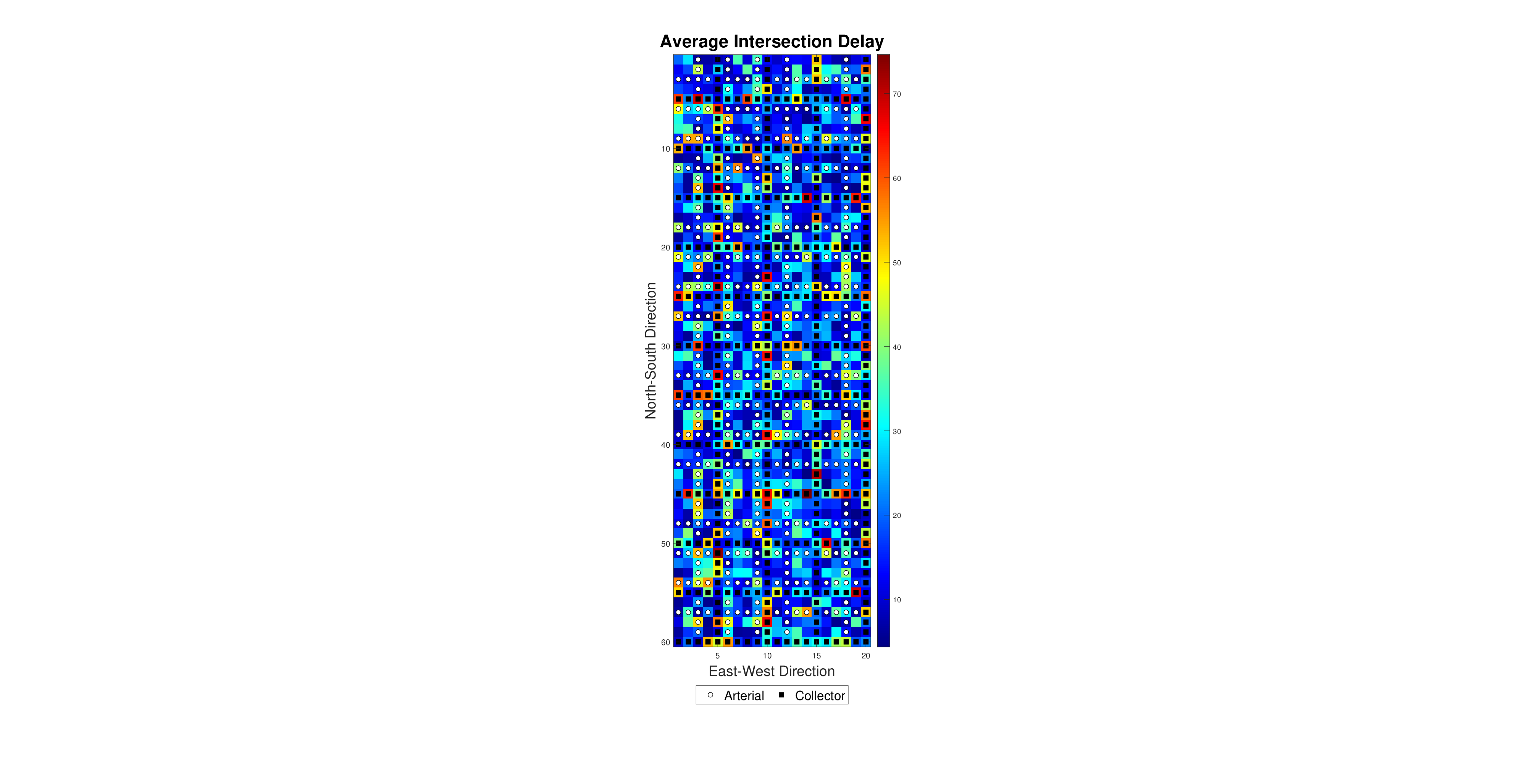}
		\caption{NSDE}
		\label{fig:p-nsde}
	\end{subfigure}
	\hfill
	\begin{subfigure}[!htbp]{0.17\linewidth}
		\centering
		\includegraphics[width=1\textwidth, trim={270mm 10mm 270mm 10mm}, clip]{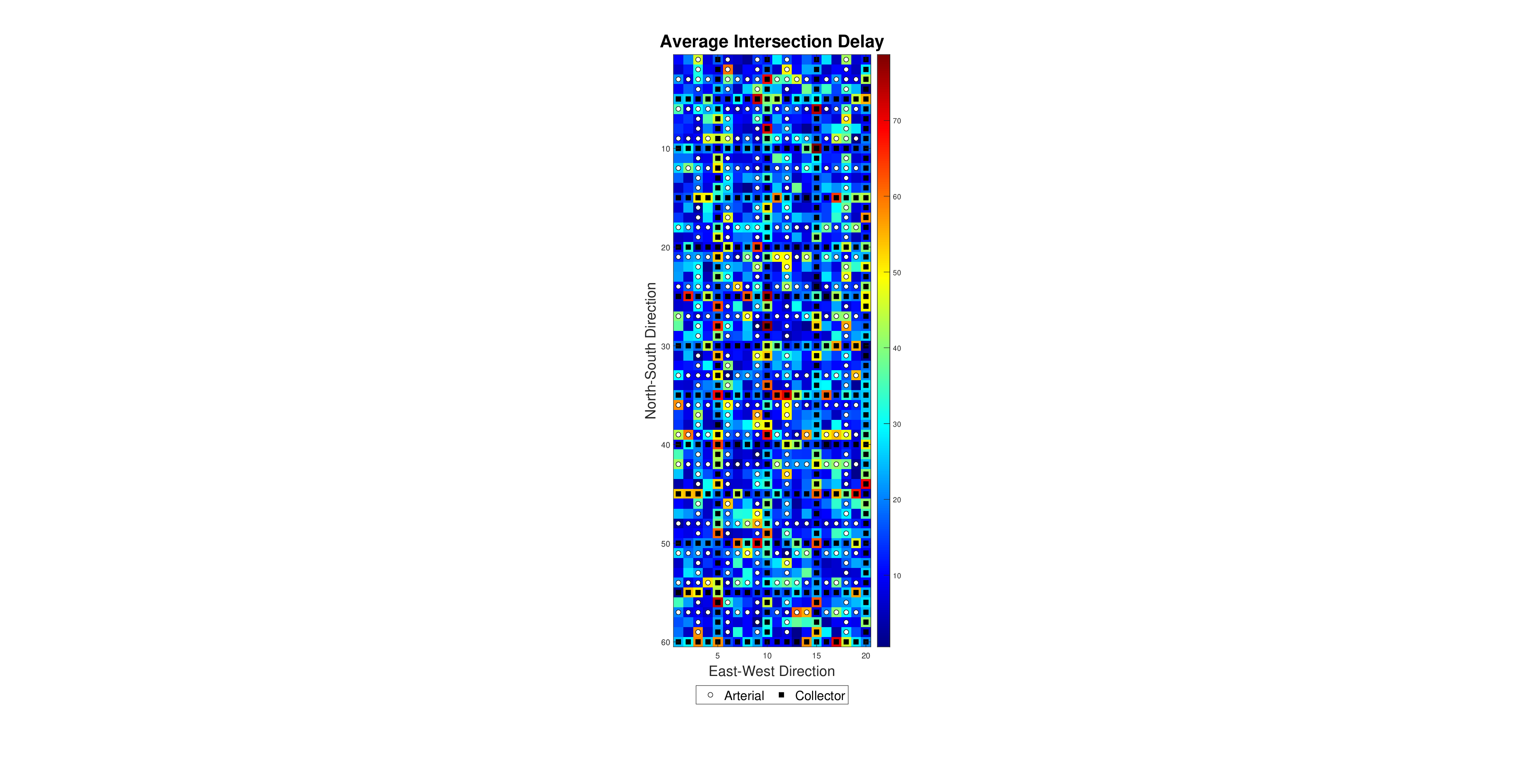}
		\caption{MOEA/D}
		\label{fig:p-moead}
	\end{subfigure}
	\hfill
	\begin{subfigure}[!htbp]{0.17\linewidth}
		\centering
		\includegraphics[width=1\textwidth, trim={270mm 10mm 270mm 10mm}, clip]{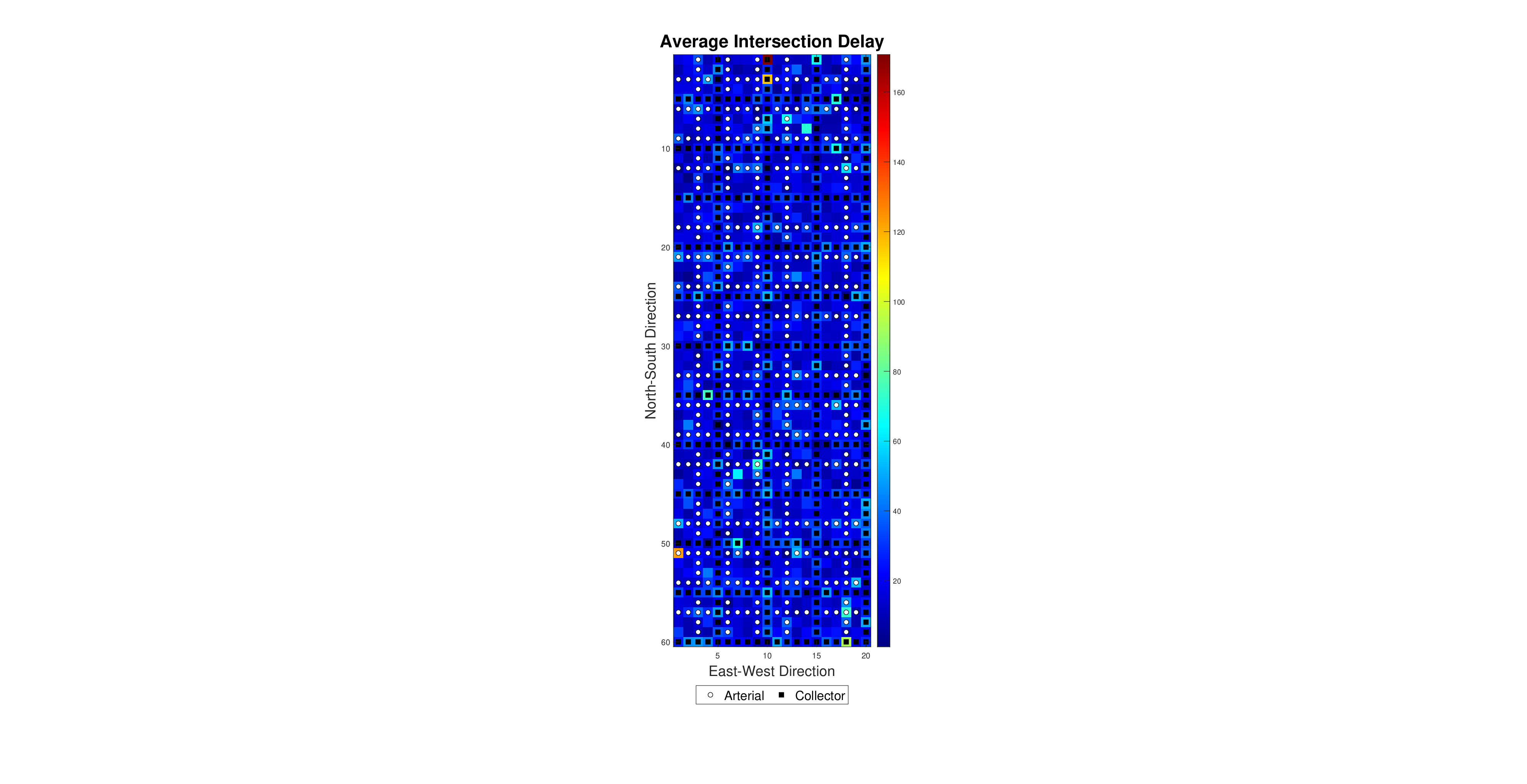}
		\caption{AHMOA}
		\label{fig:p-ahmoa}
	\end{subfigure}
	\caption{Comparison of Algorithm Performances on Average Intersection Delay in Paris City}
	\label{fig:intersectiondelay_p}
\end{figure*}

In the Paris scenario, to offer a clearer comparison of performance, we present Fig.~\ref{fig:intersectiondelay_p}, where a solution from the Pareto front of each algorithm has been selected for visualization, illustrating the average delay over a 24-hour period. Fig.~\ref{fig:p-before} displays the average delay at each intersection before optimization, showing the baseline conditions. In the optimization simulations, we modeled 20 arterial roads and 60 collector roads, resulting in a total of 1,200 intersections. The baseline scenario reveals significant red areas throughout the Paris region, indicating high average delays in the absence of optimization. Fig.~\ref{fig:m-ahmoa} illustrates the performance of the AHMOA algorithm, where the majority of the region is represented in blue, signifying minimal delays. This highlights the effectiveness of AHMOA in drastically reducing delays. In comparison to the other algorithms, the dominance of blue in AHMOA’s performance visualization further underscores its superiority in the Paris scenario. The performances of NSGA, NSDE, and MOEA/D are shown in Fig.~\ref{fig:m-nsga}, Fig.~\ref{fig:m-nsde}, and Fig.~\ref{fig:m-moead}, respectively. While these algorithms achieve some level of optimization, the presence of more red, orange, and green areas reflects their relatively suboptimal performance compared to AHMOA, with higher delays persisting in certain parts of the network. The contrasting results clearly demonstrate AHMOA’s superior ability to optimize traffic flow and minimize delays across the network, particularly in complex urban settings like Paris.

\subsubsection{Experiments on São Paulo Traffic Network}
In the São Paulo scenario, we used 30 arterial roads and 50 collectors. The global Pareto solutions are still obtained by the MOEA/D and AHMOA algorithms (shown in Table~\ref{tab:glo_sao_paulo}), with AHMOA achieving significantly more global Pareto solutions than MOEA/D. While the difference in average delay is minimal, AHMOA demonstrates superior network stability and robustness. NSGA3 and NSDE3, in comparison to the other four algorithms, did not obtain any global Pareto solutions.
\begin{table}[htbp]
	\centering
	\caption{Global Pareto Solutions Sorted by Algorithm for São Paulo}
	\begin{adjustbox}{max width=\textwidth}
		\begin{tabular}{cccc}
			\toprule
			\textbf{Algorithm} & \textbf{Log(Objective 1)} & \textbf{Log(Objective 2)} & \textbf{Log(Objective 3)} \\
			\midrule
			\multirow{14}{*}{MOEA}
			& 7.9549 & 8.0187 & -Inf \\
			& 7.9551 & 8.0183 & -7.3404 \\
			& 7.9552 & 8.0183 & -8.1186 \\
			& 7.9555 & 8.0182 & -7.5165 \\
			& 7.9556 & 8.0177 & -7.4196 \\
			& 7.9559 & 8.0146 & -Inf \\
			& 7.9565 & 8.0145 & -7.5165 \\
			& 7.9569 & 8.0141 & -7.6414 \\
			& 7.9574 & 8.0141 & -7.8175 \\
			& 7.9575 & 8.0140 & -7.4196 \\
			& 7.9577 & 8.0140 & -7.5165 \\
			& 7.9580 & 8.0138 & -7.6414 \\
			& 7.9583 & 8.0137 & -7.5165 \\
			& 7.9583 & 8.0138 & -Inf \\
			\midrule
			\multirow{23}{*}{AHMOA}
			& 7.9584 & 7.9978 & -7.3404 \\
			& 7.9585 & 7.9995 & -8.1186 \\
			& 7.9586 & 7.9993 & -7.5165 \\
			& 7.9586 & 7.9994 & -7.6414 \\
			& 7.9590 & 7.9993 & -7.6414 \\
			& 7.9591 & 7.9976 & -7.2735 \\
			& 7.9592 & 7.9975 & -7.6414 \\
			& 7.9593 & 7.9975 & -7.8175 \\
			& 7.9594 & 7.9973 & -8.1186 \\
			& 7.9597 & 7.9972 & -7.8175 \\
			& 7.9600 & 7.9967 & -7.5165 \\
			& 7.9601 & 7.9967 & -7.6414 \\
			& 7.9601 & 7.9968 & -7.8175 \\
			& 7.9603 & 7.9965 & -7.4196 \\
			& 7.9603 & 7.9966 & -7.8175 \\
			& 7.9606 & 7.9965 & -7.6414 \\
			& 7.9608 & 7.9965 & -Inf \\
			& 7.9609 & 7.9964 & -8.1186 \\
			& 7.9615 & 7.9963 & -7.4196 \\
			& 7.9615 & 7.9964 & -Inf \\
			& 7.9616 & 7.9962 & -7.6414 \\
			& 7.9617 & 7.9963 & -8.1186 \\
			& 7.9622 & 7.9962 & -7.8175 \\
			\bottomrule
		\end{tabular}
	\end{adjustbox}
	\label{tab:glo_sao_paulo}
\end{table}

\begin{figure*}[!htbp]
	\centering
	\begin{subfigure}[!htbp]{0.16\linewidth}
		\centering
		\includegraphics[width=\textwidth, trim={0mm 0mm 0mm 0mm}, clip]{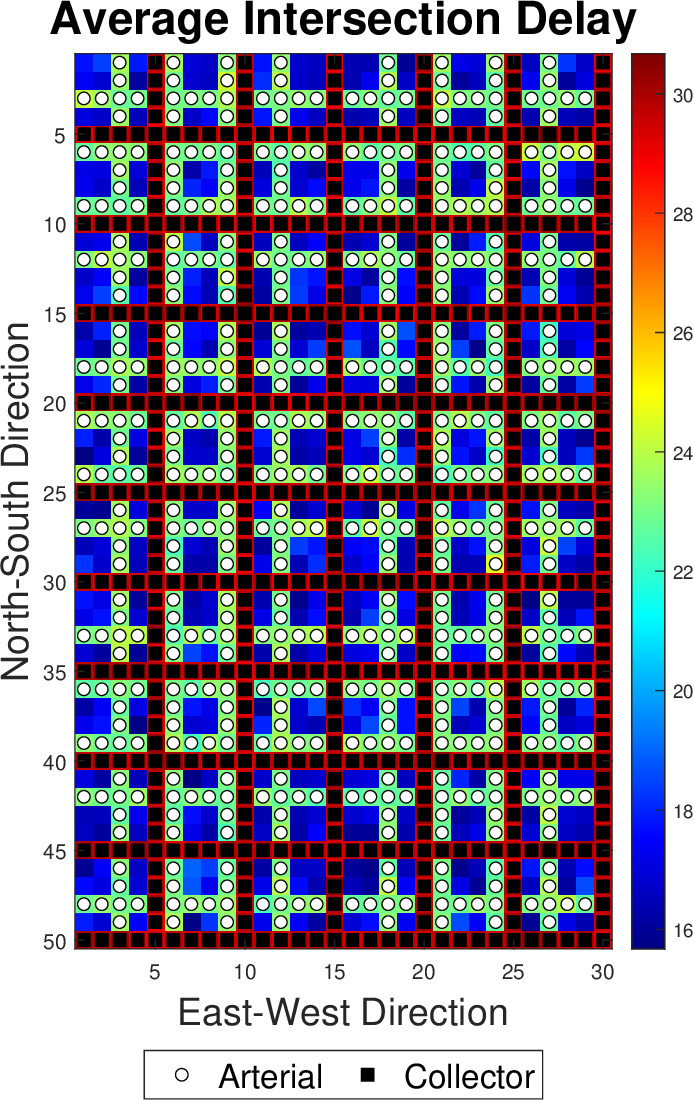}
		\caption{Baseline}
		\label{fig:s-before}
	\end{subfigure}
	\hfill
	\begin{subfigure}[!htbp]{0.16\linewidth}
		\centering
		\includegraphics[width=\textwidth, trim={0mm 0mm 0mm 0mm}, clip]{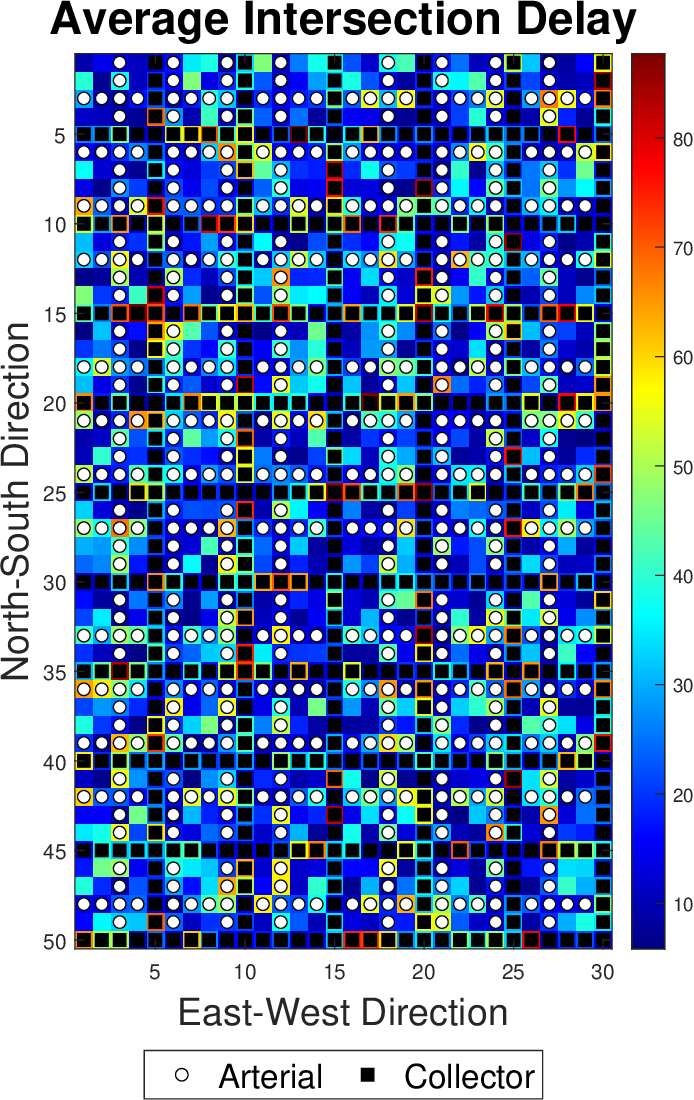}
		\caption{NSGA-III}
		\label{fig:s-nsga}
	\end{subfigure}
	\hfill
	\begin{subfigure}[!htbp]{0.16\linewidth}
		\centering
		\includegraphics[width=\textwidth, trim={0mm 0mm 0mm 0mm}, clip]{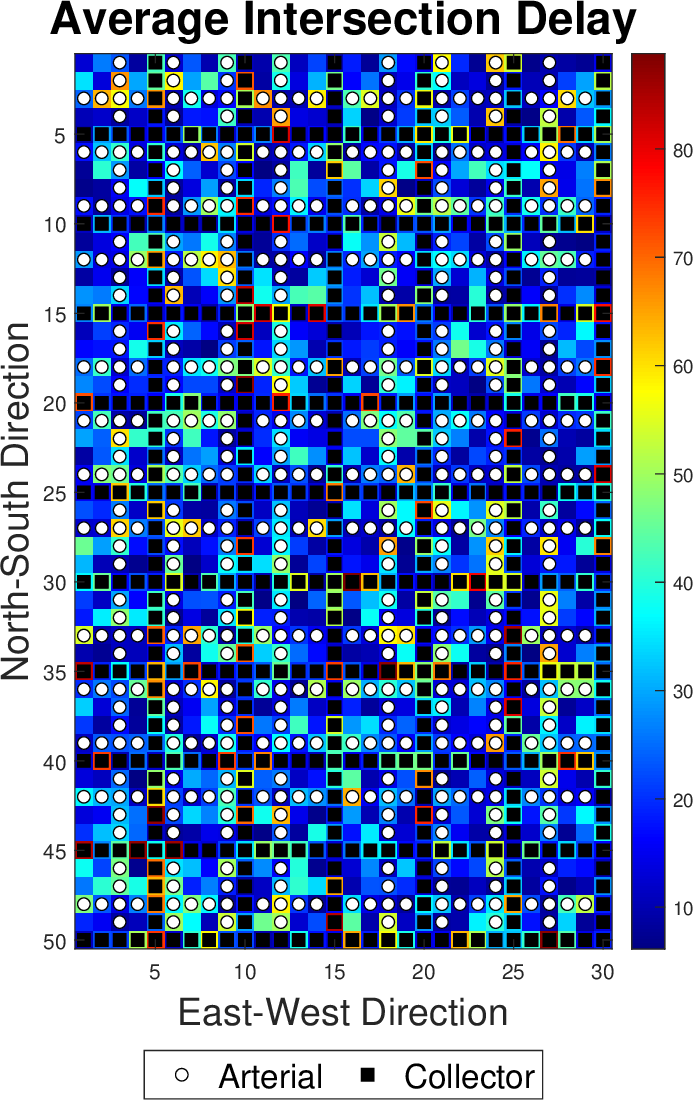}
		\caption{NSDE}
		\label{fig:s-nsde}
	\end{subfigure}
	\hfill
	\begin{subfigure}[!htbp]{0.16\linewidth}
		\centering
		\includegraphics[width=\textwidth, trim={0mm 0mm 0mm 0mm}, clip]{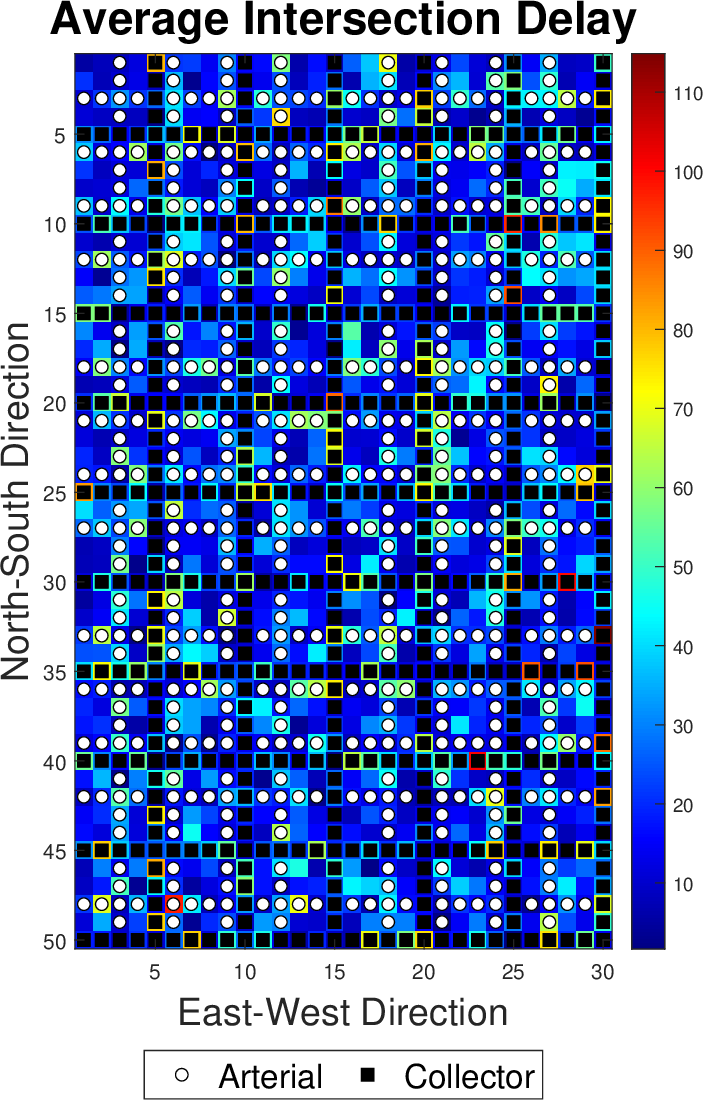}
		\caption{MOEA/D}
		\label{fig:s-moead}
	\end{subfigure}
	\hfill
	\begin{subfigure}[!htbp]{0.16\linewidth}
		\centering
		\includegraphics[width=\textwidth, trim={0mm 0mm 0mm 0mm}, clip]{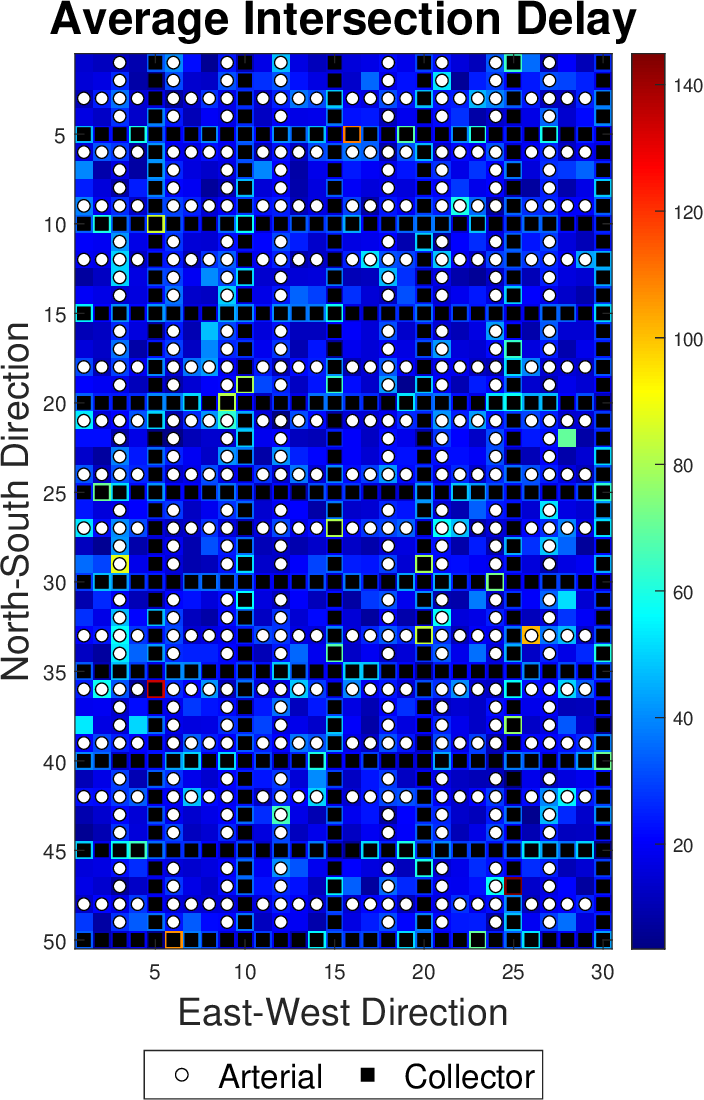}
		\caption{AHMOA}
		\label{fig:s-ahmoa}
	\end{subfigure}
	\caption{Comparison of Algorithm Performances on Average Intersection Delay in São Paulo City}
	\label{fig:intersectiondelay_s}
\end{figure*}

As shown in Fig.~\ref{fig:intersectiondelay_s}, in the São Paulo scenario, the Pareto solutions obtained by AHMOA exhibit a more prominent blue coloration overall (shown in Fig.~\ref{fig:s-ahmoa}), indicating that this algorithm is more capable of achieving lower average delays. MOEA/D follows closely, as seen in Fig.~\ref{fig:s-moead}, with a significant amount of blue regions, although there is also a noticeable presence of cyan, suggesting that some local areas still experience relatively higher average delays. In Fig.~\ref{fig:s-nsga} and Fig.~\ref{fig:s-nsde}, there are noticeably more cyan, yellow, and even red regions, indicating that the solutions obtained by NSGA3 and NSDE3 show a greater disparity when applied to São Paulo's scenario.

\subsubsection{Performance Comparison Summary}
The experimental results across all scenarios—Manhattan, Istanbul, Paris, and São Paulo—clearly demonstrate the superiority of the AHMOA algorithm compared to its peers (MOEA/D, NSGA3, and NSDE3). In every test case, AHMOA consistently identifies more global Pareto optimal solutions, reflecting its robustness in addressing large-scale, multi-objective traffic optimization problems. In the Manhattan and São Paulo scenarios, AHMOA not only outperforms on the global Pareto front but also achieves lower average delays across most intersections, demonstrating its effectiveness at minimizing traffic congestion. In Istanbul and Paris, while MOEA/D performs well in certain aspects, AHMOA excels in network stability and robustness, providing a more diverse and well-distributed set of solutions.

Therefore, we can conclude that AHMOA's ability to balance multiple objectives while maintaining robustness under dynamic traffic conditions makes it the most comprehensive and effective algorithm across all tested traffic networks. The consistent performance of AHMOA in reducing delays, optimizing traffic flow, and providing stable solutions underscores its suitability for large-scale urban traffic systems.

\section{Conclusions and Future Work}
\label{sec:con}
This paper introduces a novel large-scale multi-objective optimization approach for robust traffic signal control in uncertain urban environments, representing a significant advancement in urban traffic management. The proposed algorithm effectively addresses the complexities of modern urban traffic networks, successfully balancing multiple conflicting objectives while maintaining system stability under unpredictable traffic conditions. The robustness and versatility of the approach have been validated through comprehensive simulations in four major cities—Manhattan, Paris, São Paulo, and Istanbul—each of which presents distinct challenges in network topology, traffic dynamics, and operational uncertainties. Despite these variations, the algorithm consistently delivered reliable performance across all scenarios, demonstrating its potential to transform traffic management strategies in diverse urban contexts globally.

A key strength of our method is its scalability to large-scale, city-wide networks. By efficiently managing thousands of intersections, the algorithm optimizes critical factors such as minimizing travel time, maximizing throughput, and ensuring network stability. Its ability to adapt to different urban layouts and rapidly changing traffic conditions further underscores its position as a cutting-edge solution for intelligent traffic management. Additionally, this work fills a critical gap in current traffic control systems by incorporating robust performance measures that ensure reliability under uncertainty. This is particularly important in real-world applications, where traffic conditions can shift unpredictably. Our algorithm's adaptive coordination of traffic signals leads to significant improvements in overall traffic flow and congestion reduction, even in the most challenging environments.

The practical implications of this research are far-reaching. The proposed algorithm provides city planners and traffic engineers with a powerful tool for advancing smart city initiatives and promoting sustainable urban mobility. Its ability to handle large-scale networks, optimize multiple objectives, and maintain robustness in uncertain conditions places this work at the forefront of intelligent urban management systems. Future research can build upon this work by focusing on real-time implementation and testing in live urban settings, integration with emerging technologies such as connected vehicles and IoT, and refining the algorithm to address the unique challenges posed by emerging megacities. As urban populations grow and infrastructure becomes increasingly complex, the need for adaptive and robust traffic management systems becomes critical. This study represents a significant step toward meeting these demands, contributing to the development of more efficient, sustainable, and livable cities.

\bibliographystyle{IEEEtran}
\bibliography{references.bib}

\vspace{0pt}
\begin{IEEEbiography}[{\includegraphics[width=1in,height=1.25in,clip,keepaspectratio]{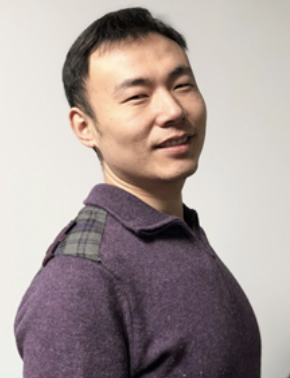}}]{Weian GUO} received the M.Eng. degree in navigation, guidance, and control from Northeastern University, Shenyang, China, in 2009, and the doctor of engineering degree from Tongji University, Shanghai, China, in 2014. From 2011 to 2013, he was sponsored by China Scholarship Council to carry on his research at the Social Robotics Laboratory, National University of Singapore. He is currently an associate Professor with the Sino-German College of Applied Science, Tongji University. His interests include computational intelligence, optimization, artificial intelligence, and control theory.
\end{IEEEbiography}
\begin{IEEEbiography}[{\includegraphics[width=1in,height=1.25in,clip,keepaspectratio]{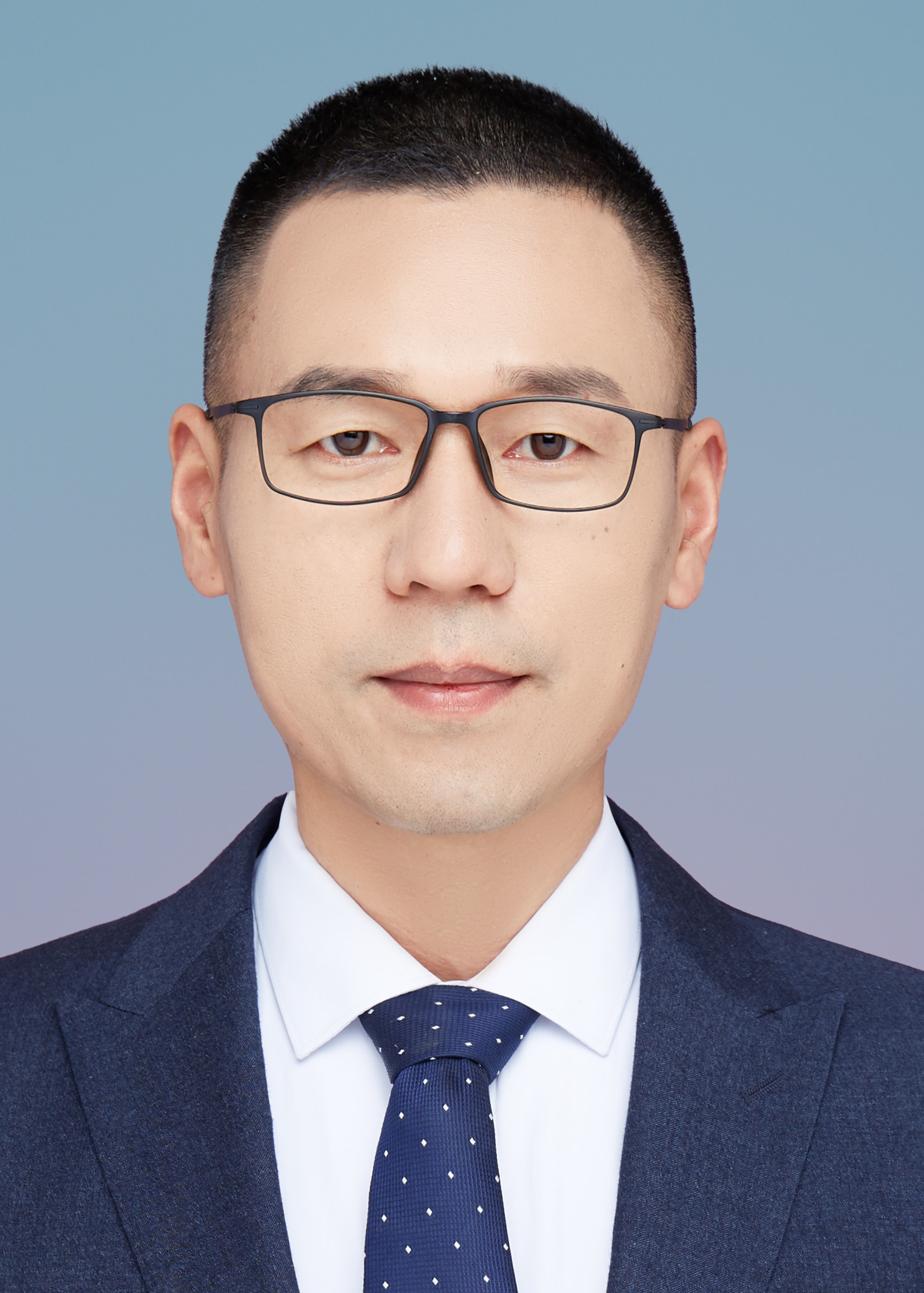}}]{Wuzhao LI} received the Ph.D. degree in Control Science and Engineering from Tongji University, Shanghai, China. In 2022, he was a Professor with the Wenzhou Polytechnic, China. He has authored or coauthored more than 20 papers in international SCI journals. His research interests include machine learning, Multi-Objective optimization and robust control.
\end{IEEEbiography}
\begin{IEEEbiography}[{\includegraphics[width=1in,height=1.25in,clip,keepaspectratio]{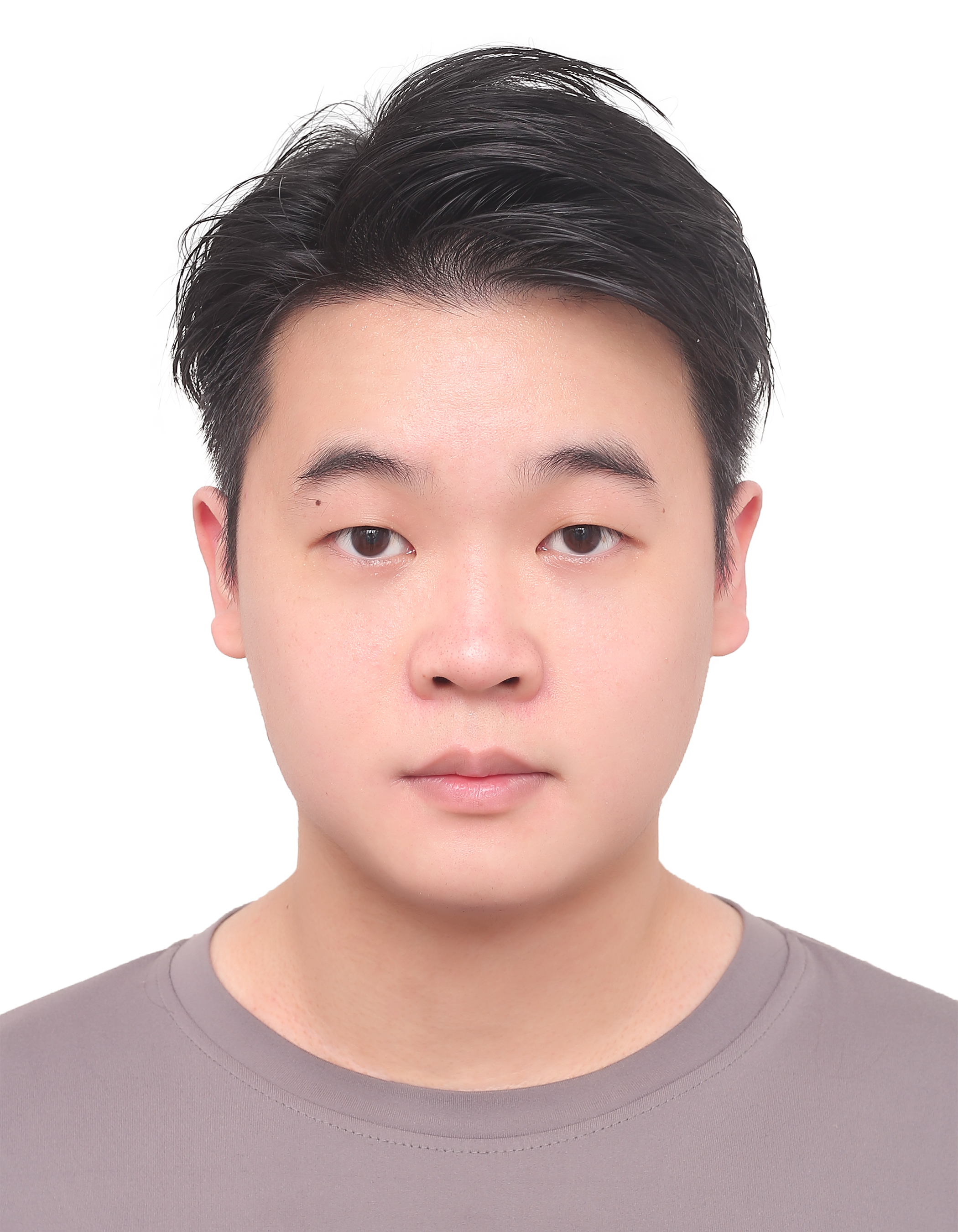}}]{Zhiou ZHANG} received bachelor degree in vehicle service engineering from Tongji University, Shanghai, China. In 2024, he is a postgraduate of information technology with University of New South Wales, Sydney, Australia. His research interests include data management, vehicle vision and control system optimization.
\end{IEEEbiography}
\begin{IEEEbiography}[{\includegraphics[width=1in,height=1.25in,clip,keepaspectratio]{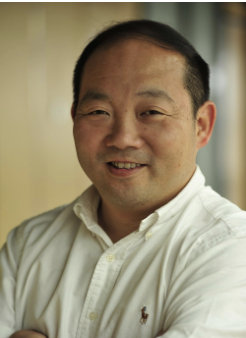}}]{Lun ZHANG}
	received the B.S. and Ph.D degrees in computer communications, transportation information engineering and control, Tongji University, Shanghai, China, in 1992 and 2005, respectively. He is currently a professor with School of Transportation, Tongji University. His interests include intelligent transportation, computational intelligence, and deep learning.
\end{IEEEbiography}	
\begin{IEEEbiography}[{\includegraphics[width=1in,height=1.25in,clip,keepaspectratio]{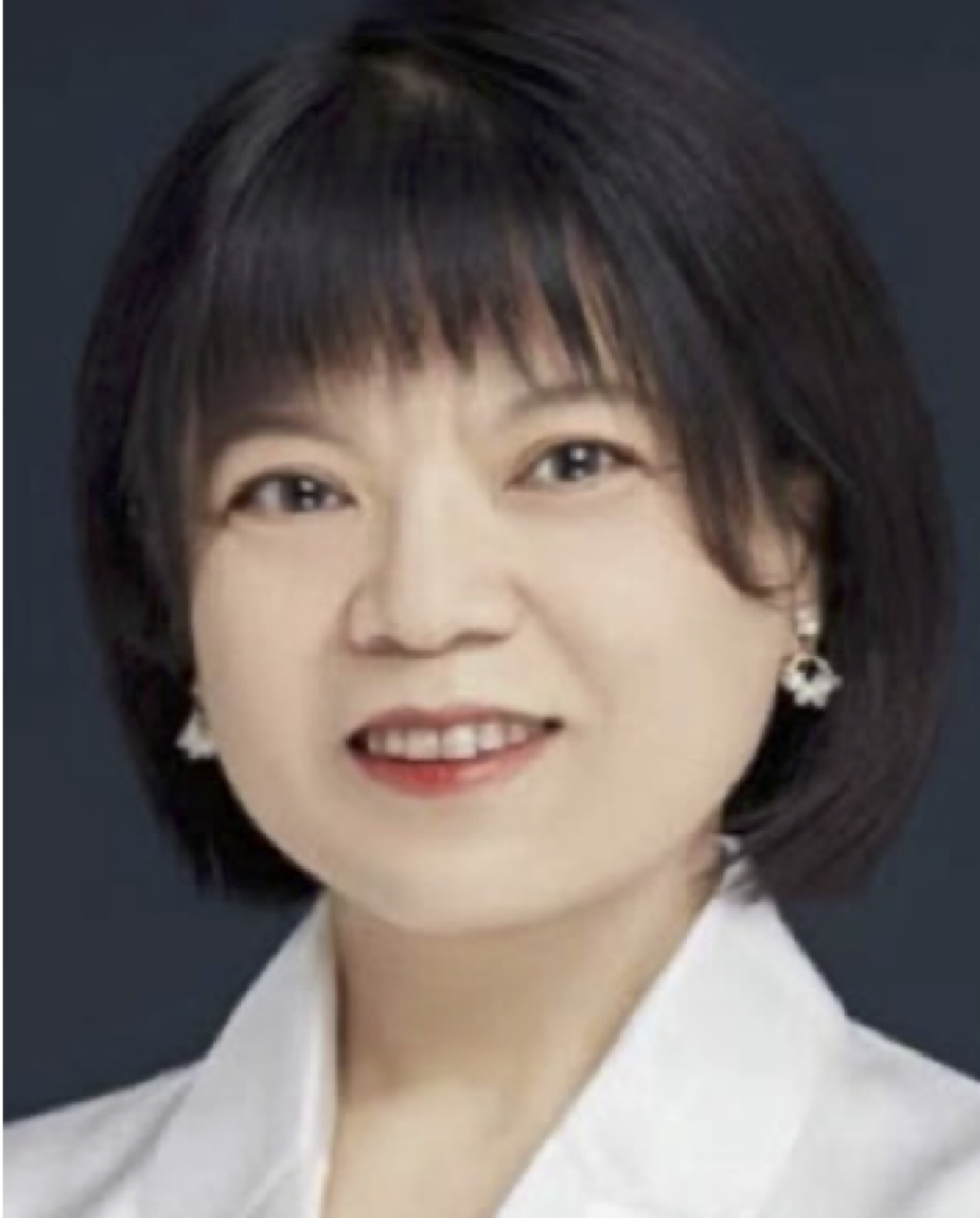}}]{Li LI} received the B.S. and M.S. degrees in electrical automation from Shengyang Agriculture University, Shengyang, China, in 1996 and 1999, respectively, and the Ph.D. degree in mechatronics engineering from the Shenyang Institute of Automation, Chinese Academy of Science, Shenyang, in 2003. She joined Tongji University, Shanghai, China, in 2003, where she is currently a professor of control science and engineering. She has over 50 publications, including 4 books, over 30 journal papers, and 2 book chapters. Her current research interests include production planning and scheduling, computational intelligence, data-driven modeling and optimization, semiconductor manufacturing, and energy systems.
\end{IEEEbiography}
\begin{IEEEbiography}[{\includegraphics[width=1in,height=1.25in,clip,keepaspectratio]{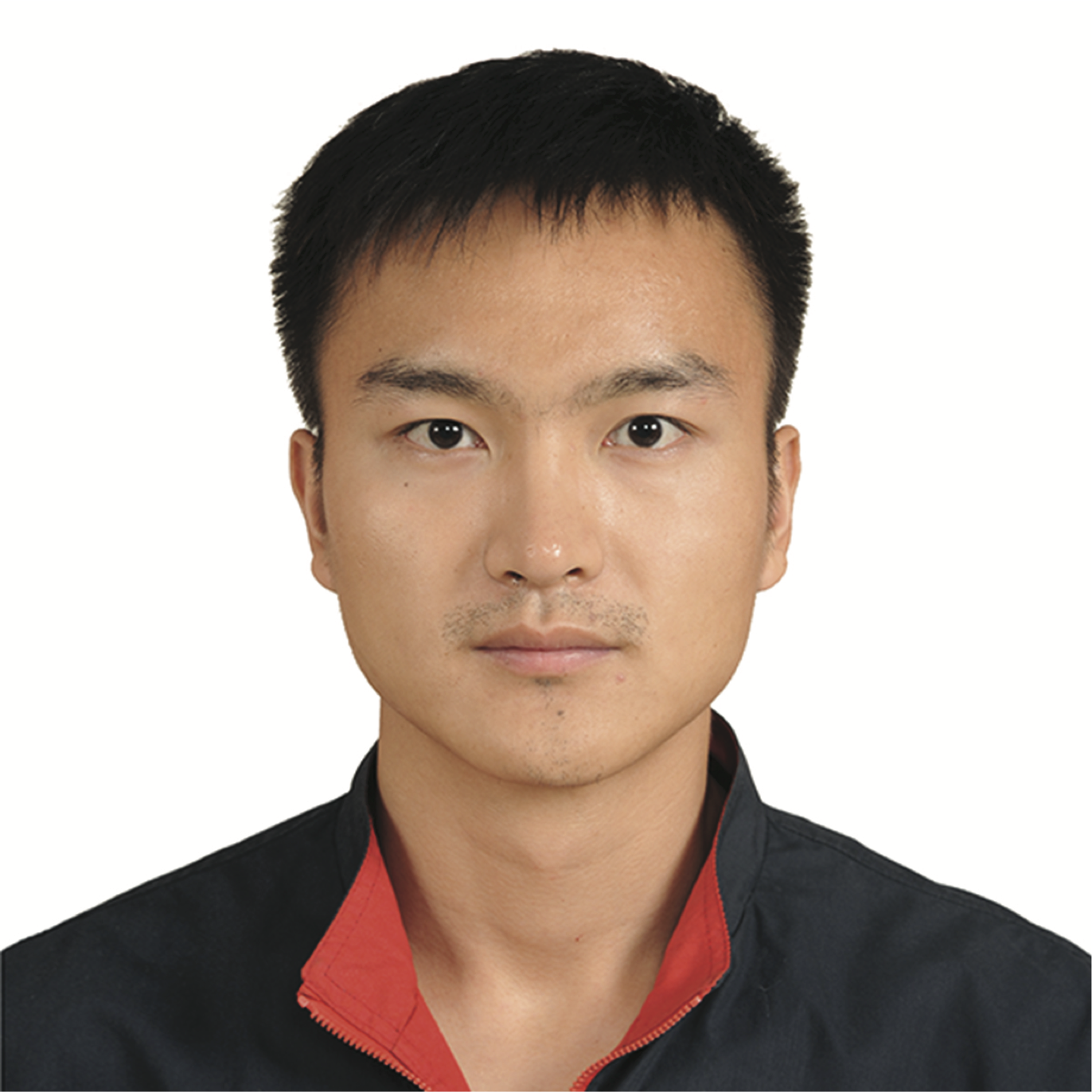}}]{Dongyang Li}
received the M.S. and Ph.D degrees in school of electronics and information engineering, Tongji University, Shanghai, China, in 2017 and 2022, respectively. From 2019 to 2021, he was sponsored by China Scholarship Council to carry on his research at Georgia Institute of Technology. He is now an engineer with the Sino-German College of Applied Science, Tongji University; His research interest includes computational intelligence, deep learning and their applications.
\end{IEEEbiography}

\end{document}